\documentclass[a4paper,fleqn]{cas-dc}

\usepackage[numbers]{natbib}
\usepackage[ruled, vlined, linesnumbered]{algorithm2e}
\usepackage{subcaption}
\usepackage{amsmath,amssymb}
\usepackage{multirow}
\usepackage{enumerate}

\begin{document}
\let\WriteBookmarks\relax
\def\floatpagepagefraction{1}
\def\textpagefraction{.001}
\shorttitle{Robust Representation Learning for Spiking Neural Networks based on Conditional Information Bottleneck}
\shortauthors{Y. Zha et~al.}

\title [mode = title]{Robust Representation Learning for Spiking Neural Networks based on Conditional Information Bottleneck}                      

\author[1]{Yuli Zha}
\credit{Conceptualization, Methodology, Validation, Software, Investigation, Writing -- Original draft}

\author[1]{Wang Fan}
\credit{Software, Validation}

\author[1]{Yixiao Dai}
\credit{Software, Data curation}

\author[1]{Lingjie Wang}
\credit{Methodology, Investigation}

\author[1]{Bang Hu}
\credit{Investigation, Validation}

\author[2]{Wei Cao}
\credit{Methodology, Writing -- Review \& Editing}

\author[2]{Fengzhe Zhang}
\credit{Conceptualization, Methodology}

\author[2]{Fan Zhang}[orcid=0000-0001-7456-8377]
\cormark[1]
\ead{ffzhang@fudan.edu.cn}
\credit{Conceptualization, Resources, Supervision, Writing -- Review \& Editing}

\affiliation[1]{organization={College of Computer Science and Artificial Intelligence, Fudan University},
                city={Shanghai},
                postcode={200433},
                country={China}}

\affiliation[2]{organization={Institute of Big Data, Fudan University},
                city={Shanghai},
                postcode={200433},
                country={China}}
                
\cortext[1]{Corresponding author}

\begin{abstract}
Spiking Neural Networks (SNNs) have shown great potential for low-power neuromorphic computing due to their event-driven processing and spatiotemporal sparsity, but they remain vulnerable to input perturbations, hardware-intrinsic noise, and adversarial attacks in real-world deployment. We attribute this vulnerability partly to input-specific hidden representations learned by standard SNNs trained with conventional cross-entropy loss, as our information-plane analysis empirically shows that substantial input-related information can still be retained after label-predictive information becomes stable. To address this issue, we propose SCIB, a simple yet effective robust learning objective for SNNs based on the Conditional Information Bottleneck. Different from conventional variational IB methods, SCIB replaces the mutual information compression term $I(X;S)$ with the conditional mutual information $I(X;S|Y)$ and derives a tractable variational upper bound. Moreover, it introduces a learnable class-conditional prior instead of a label-agnostic isotropic Gaussian prior, improving the expressiveness of variational compression and encouraging class-structured representations when combined with the supervised classification objective. Systematic experiments on Fashion-MNIST, CIFAR-10, and CIFAR-100 under external environmental noise, neuromorphic hardware noise, and adversarial attacks show that SCIB improves robustness over standard SNNs and IB-based baselines under most evaluated noise conditions, improves adversarial accuracy under the evaluated attacks without using adversarial training, and reduces the SynOps-based single-sample inference energy consumption by 24.9\% relative to an unconstrained SNN baseline.
\end{abstract}


\maketitle

\section{Introduction}
Neuromorphic computing, which emulates the operating mechanisms of the biological brain, offers a promising pathway for overcoming the energy-efficiency bottleneck of conventional von Neumann architectures~\cite{Brain2022Mehonic, System2020Zhang}. As the core computational paradigm of neuromorphic computing, Spiking Neural Networks (SNNs) encode and transmit information through discrete spike events, and are characterized by event-driven processing, temporal sensitivity, and inherent sparsity~\cite{Networks1997Maass, Towards2019Roy}. On neuromorphic hardware platforms such as Intel Loihi~\cite{Loihi2018Davies}, IBM TrueNorth~\cite{Truenorth2015Akopyan}, and Tianjic~\cite{Tianjic2020Deng}, SNNs can achieve remarkable energy-efficiency advantages over conventional architectures~\cite{Advancing2021Davies, Opportunities2022Schuman}. In recent years, surrogate gradient-based direct training methods~\cite{Direct2019Wu, Surrogate2019Neftci} have effectively alleviated the non-differentiability of the spike firing function, promoting the application of SNNs to image classification~\cite{Spatio2018Wu, Spiking2020Kim}, object detection~\cite{Spiking2015Cao}, and other neuromorphic perception tasks.

However, SNNs still face severe robustness challenges in real-world deployment environments. The perturbations encountered by SNNs can be broadly categorized into three types. First, environmental perturbations and random input noise directly corrupt spike-encoded signals at the sensory or input-encoding stage~\cite{Comprehensive2019Sharmin}. Second, non-ideal factors intrinsic to neuromorphic hardware, such as thermal noise, synaptic weight drift, and stochastic membrane-potential fluctuations, may introduce deviations in spike timing and neuronal responses~\cite{BiCoSS2021Yang}. Third, surrogate gradient-based adversarial attacks can exploit differentiable approximations of spike dynamics to construct malicious perturbations, posing serious threats to safety-critical SNN applications~\cite{Exploring2021Liang, Rate2023Bu}.

Existing studies have developed different robustness-enhancement strategies for these perturbation sources. For input-level noise, methods including spike encoding optimization, crafted-noise training, and frequency-domain filtering aim to suppress perturbations before or during spike generation~\cite{Inherent2020Sharmin, Hire2021Kundu, FEEL2024Xu}. For hardware-induced neuronal perturbations, bio-inspired stochastic mechanisms and heterogeneous neuronal dynamics improve robustness by injecting structured randomness into spike transmission, membrane evolution, or synaptic dynamics~\cite{Enhancing2024Ding, Random2025Wang, Noise2025Garipova}. For gradient-based adversarial perturbations, regularization-based and adversarial training methods constrain perturbation propagation by imposing spectral, Lipschitz-type, or gradient-level restrictions on SNNs~\cite{Snn2022Ding, Regularization2023Zhang, Enhancing2024Liu}. Although these methods improve SNN robustness from different perspectives, they are often designed for specific perturbation models or attack types. Encoding-based methods mainly operate near the input stage, stochastic mechanisms may require carefully designed noise processes or modified neuron models, and adversarial training usually relies on attack-specific sample generation and expensive inner-loop optimization. For these reasons, it is desirable to develop a simple objective-level surrogate for SNN training that is easy to implement, does not alter the SNN architecture or neuronal dynamics, and can generalize across heterogeneous perturbation sources.

Our approach is motivated by an empirical information-plane observation: trained SNN representations can retain substantial or even redundant input-related information even after their label-predictive information becomes stable~\cite{Opening2017Shwartz, On2021Geiger}. This observation suggests that spike representations may contain input-specific variations, some of which can be perturbation-sensitive or redundant for the target classification task and may amplify sensitivity to environmental noise, hardware perturbations, and adversarial attacks.

Information Bottleneck (IB) theory provides a principled framework for learning minimally sufficient representations by preserving label-relevant information while compressing input-dependent redundancy~\cite{Information2000Tishby, Deep2015Tishby}. It has been widely applied to conventional neural networks, including variational deep networks, convolutional neural networks, recurrent or sequential models, and robust representation learning frameworks, to improve compactness, generalization, and adversarial robustness~\cite{Deep2016Alemi, Nonlinear2019Kolchinsky, How2023Kawaguchi, Improving2023Kuang, Supervised2025Weingarten, Gated2023Alesiani}. However, compared with its extensive use in conventional deep neural networks, the systematic use of IB principles for improving SNN robustness remains relatively underexplored.

Unfortunately, directly applying conventional IB objective or the Variational Information Bottleneck (VIB)~\cite{Deep2016Alemi} to SNNs may lead to suboptimal performance. The first difficulty lies in the compression target. Conventional IB objectives typically penalize the mutual information $I(X;S)$ between the input $X$ and the learned representation $S$. Under the supervised representation-learning Markov chain $Y \leftrightarrow X \leftrightarrow S$, where $S$ is generated from $X$ and has no additional access to $Y$ at inference time, the total information $I(X;S)$ can be decomposed into the label-relevant predictive term $I(S;Y)$ and the label-conditioned term $I(X;S|Y)$. Therefore, directly penalizing $I(X;S)$ may compress useful discriminative information together with nuisance factors, potentially leading to a robustness--accuracy trade-off.~\cite{Conditional2020Fischer}. The second difficulty lies in the variational prior. Standard VIB usually adopts a label-agnostic unimodal Gaussian prior, such as $\mathcal{N}(0,I)$, which pulls representations from all classes toward a single distributional anchor. Such a prior has limited expressiveness for class-structured spike representations and may weaken inter-class separability, especially when the number of categories increases or when noisy spike features are involved. These limitations motivate a more selective compression objective and a class-aware prior structure for robust SNN representation learning.

To address the above limitations, we propose SCIB, a Spiking Conditional Information Bottleneck framework for robust SNNs. Instead of using the conventional mutual information $I(X;S)$ as the compression objective, SCIB takes the conditional mutual information $I(X;S|Y)$ as the ideal compression target and implements it through a tractable variational upper-bound regularizer. We further introduce a learnable class-conditional prior, where each class is assigned an independent distributional anchor in the latent space. This design provides a more expressive prior than the label-agnostic standard Gaussian used in conventional VIB, while encouraging intra-class compactness and helping maintain inter-class separability when combined with the classification objective. SCIB can be integrated into standard surrogate-gradient training as an objective-level regularizer, without modifying the SNN architecture or neuronal dynamics. The main contributions of this paper are summarized as follows:

\begin{itemize}

\item Conditional information bottleneck framework for robust SNNs. Unlike perturbation-specific defense methods and conventional IB objectives based on indiscriminate compression of $I(X;S)$, SCIB is motivated by $I(X;S|Y)$ and implements conditional compression through a tractable variational upper-bound regularizer. This design targets label-conditioned input-specific information while relying on the supervised loss to preserve predictive information, thereby mitigating the robustness--accuracy tension observed with indiscriminate IB compression.

\item Learnable class-conditional prior. We derive a tractable variational upper bound for the conditional compression term and instantiate it with a learnable class-conditional prior. The resulting prior can be interpreted as a Gaussian mixture over labels, providing class-specific distributional anchors for spike representations and improving the expressiveness of variational compression.

\item Systematic robustness evaluation under heterogeneous perturbations. On Fashion-MNIST, CIFAR-10, and CIFAR-100, we evaluate SCIB under environmental noise, neuromorphic hardware noise, and adversarial attacks. Results show that SCIB improves robustness over standard SNNs and existing IB-based baselines under diverse noise conditions, and improves adversarial accuracy under the evaluated attacks without adversarial training.

\item Joint robustness and energy analysis. SynOps-based analysis shows that SCIB reduces spike firing activity while improving robustness in the evaluated settings, lowering the analytical single-sample computation-energy by 24.9\% relative to the unconstrained SNN baseline.

\end{itemize}

\section{Related work}
\subsection{Robustness of spiking neural networks}
Spiking Neural Networks (SNNs) encode and transmit information through discrete spike events, where neuronal states evolve over time according to membrane-potential dynamics. This event-driven computation paradigm provides SNNs with temporal sparsity and high energy efficiency, but the non-differentiability of the spike firing function makes their training more challenging than that of conventional artificial neural networks (ANNs)~\cite{Networks1997Maass,Towards2019Roy,Supervised2017Mostafa}. Existing training paradigms mainly include ANN-to-SNN conversion~\cite{Diet2021Rathi}, membrane-potential-based learning~\cite{Rmp2020Han}, and surrogate gradient-based direct training~\cite{Direct2019Wu,Surrogate2019Neftci}. Among them, surrogate gradient learning has become the dominant approach for training deep SNNs, but the differentiable approximations introduced during training also make SNNs susceptible to gradient-based adversarial attacks~\cite{Exploring2021Liang,Rate2023Bu}.

Despite recent progress in classification accuracy and deployment efficiency, SNNs remain vulnerable to heterogeneous perturbations in real-world environments, including input-level sensory noise, hardware-induced neuronal or synaptic noise, and gradient-based adversarial perturbations~\cite{Comprehensive2019Sharmin,BiCoSS2021Yang,Exploring2021Liang,Rate2023Bu}. These perturbations can disturb spike timing, membrane-potential evolution, or input spike patterns, and may accumulate across timesteps and layers in deep SNNs. 

Existing SNN robustness methods can be roughly grouped according to the perturbation source they target. For input-level perturbations, spike encoding optimization methods aim to suppress noise at the input or temporal encoding stage. Discrete input encoding has been shown to provide partial intrinsic robustness~\cite{Inherent2020Sharmin}, while HIRE-SNN~\cite{Hire2021Kundu} and FEEL-SNN~\cite{FEEL2024Xu} further improve robustness through crafted input noise or frequency-domain encoding. For hardware-induced perturbations, stochastic modeling and bio-inspired mechanisms enhance robustness by injecting structured randomness into spike transmission, membrane evolution, or synaptic dynamics. Representative examples include stochastic gating mechanisms~\cite{Enhancing2024Ding}, random heterogeneous neuronal dynamics~\cite{Random2025Wang}, and dynamic synaptic noise modeling~\cite{Noise2025Garipova}. For adversarial perturbations, regularization-based and adversarial training methods constrain perturbation propagation by imposing explicit smoothness, spectral, or gradient-level restrictions. For example, SNN-RAT introduces Parseval regularization and mixed adversarial training to limit adversarial amplification across timesteps~\cite{Snn2022Ding}, while sparse-gradient methods improve robustness by suppressing adversarial gradient propagation~\cite{Enhancing2024Liu}.

Although these methods improve SNN robustness from different perspectives, most of them are closely tied to specific perturbation models. This leaves room for representation-level learning objectives that can suppress perturbation-sensitive redundancy without relying on a particular noise model, attack procedure, or architectural modification.

\subsection{Information bottleneck in deep neural networks}
Information Bottleneck (IB) theory provides a principled framework for learning compact and predictive representations~\cite{Information2000Tishby,Deep2015Tishby}. We denote the input by $X$, the class label by $Y$, and the learned representation by $S$. The mutual information $I(X;S)$ measures how much information about the input is retained in the representation, while $I(S;Y)$ measures how much label-relevant predictive information is preserved by the representation. For two continuous random variables $X$ and $Y$ with joint probability density function $f(x,y)$ and marginal probability density functions $f(x)$ and $f(y)$ respectively, mutual information is defined as:
\begin{equation}
    I(X;Y)=\iint f(x,y)\log\frac{f(x,y)}{f(x)f(y)}\,dx\,dy.
    \label{eq:mi}
\end{equation}

Mutual information measures the strength of statistical dependence between $X$ and $Y$, with $I(X;Y)=0$ if and only if $X$ and $Y$ are statistically independent. From the perspective of information entropy, mutual information can be equivalently decomposed as:
\begin{equation}
    I(X;Y)=H(X)+H(Y)-H(X,Y)=H(X)-H(X|Y),
    \label{eq:mi_decomp}
\end{equation}
where $H(\cdot)$, $H(\cdot,\cdot)$, and $H(\cdot|\cdot)$ denote Shannon entropy, joint entropy, and conditional entropy, respectively. Eq.~\eqref{eq:mi_decomp} reveals the intuitive meaning of mutual information: $I(X;Y)$ equals the reduction in uncertainty about $X$ after observing $Y$.

The conventional IB objective, which is typically maximized, encourages $S$ to preserve label-relevant information while compressing input-dependent information:
\begin{equation}
\mathcal{J}_{\mathrm{IB}}
= \max I(S;Y)-\beta I(X;S),
\label{eq:ib_objective}
\end{equation}
where $\beta$ controls the trade-off between predictive sufficiency and information compression.

The IB principle has been widely studied in deep neural networks (DNNs). Variational Information Bottleneck (VIB)~\cite{Deep2016Alemi} introduces variational inference to make the IB objective tractable in high-dimensional neural networks. Nonlinear IB~\cite{Nonlinear2019Kolchinsky}, supervised IB~\cite{Supervised2025Weingarten}, information bottleneck distillation~\cite{Improving2023Kuang}, and gated IB~\cite{Gated2023Alesiani} further extend the IB principle to generalization, adversarial robustness, distillation, and sequential decision-making. Recent studies have also adapted IB objectives and information estimators to specific learning tasks. Zhang et al.~\cite{Information2025Zhang} formulated feature selection as a masked deterministic IB problem and employed matrix-based R\'enyi's $\alpha$-order entropy to obtain a tractable objective without variational approximation. Tang and Yu~\cite{InfoDPCCA2025Tang} employed variational IB and conditional mutual-information regularization to extract shared representations from two interdependent sequences. Information-plane studies also provide empirical tools for analyzing representation dynamics during neural network training~\cite{Opening2017Shwartz,On2021Geiger,Information2021Li,Adaptive2019Chelombiev}.

Compared with its extensive use in conventional DNNs, the application of IB theory to SNNs is still limited. SNIB~\cite{SNIB2023Yang} applies nonlinear information bottleneck regularization to spike-based learning and demonstrates that constraining input-representation dependence can improve SNN robustness. SeLHIB~\cite{Self2024Yang} extends high-order information bottleneck learning to self-supervised event-based optical flow estimation. These studies validate the feasibility of information compression in the spike domain, but existing IB-based SNN methods mainly inherit the conventional compression form from DNNs and have not systematically addressed conditional redundancy in spike representations or class-structured variational priors.

\section{Empirical observation}
\label{sec:empirical_observation}
Before introducing the SCIB framework, we first analyze the representation dynamics of conventionally trained SNNs and IB-regularized SNNs from the perspective of the information plane~\cite{Opening2017Shwartz,On2021Geiger}. The information plane uses the mutual information between the input and the representation, $I(X;S)$, as the horizontal axis, and the mutual information between the representation and the label, $I(S;Y)$, as the vertical axis. This visualization provides an empirical view for examining whether a network mainly fits label-relevant information or also reduces input-dependent information during training. An increase in $I(S;Y)$ indicates that the learned representation becomes more predictive of the class label, whereas a decrease in $I(X;S)$ suggests a reduction of input-related information retained.

\begin{figure*}[t]
    \centering
    \begin{subfigure}[b]{0.45\textwidth}
        \centering
        \includegraphics[width=\linewidth]{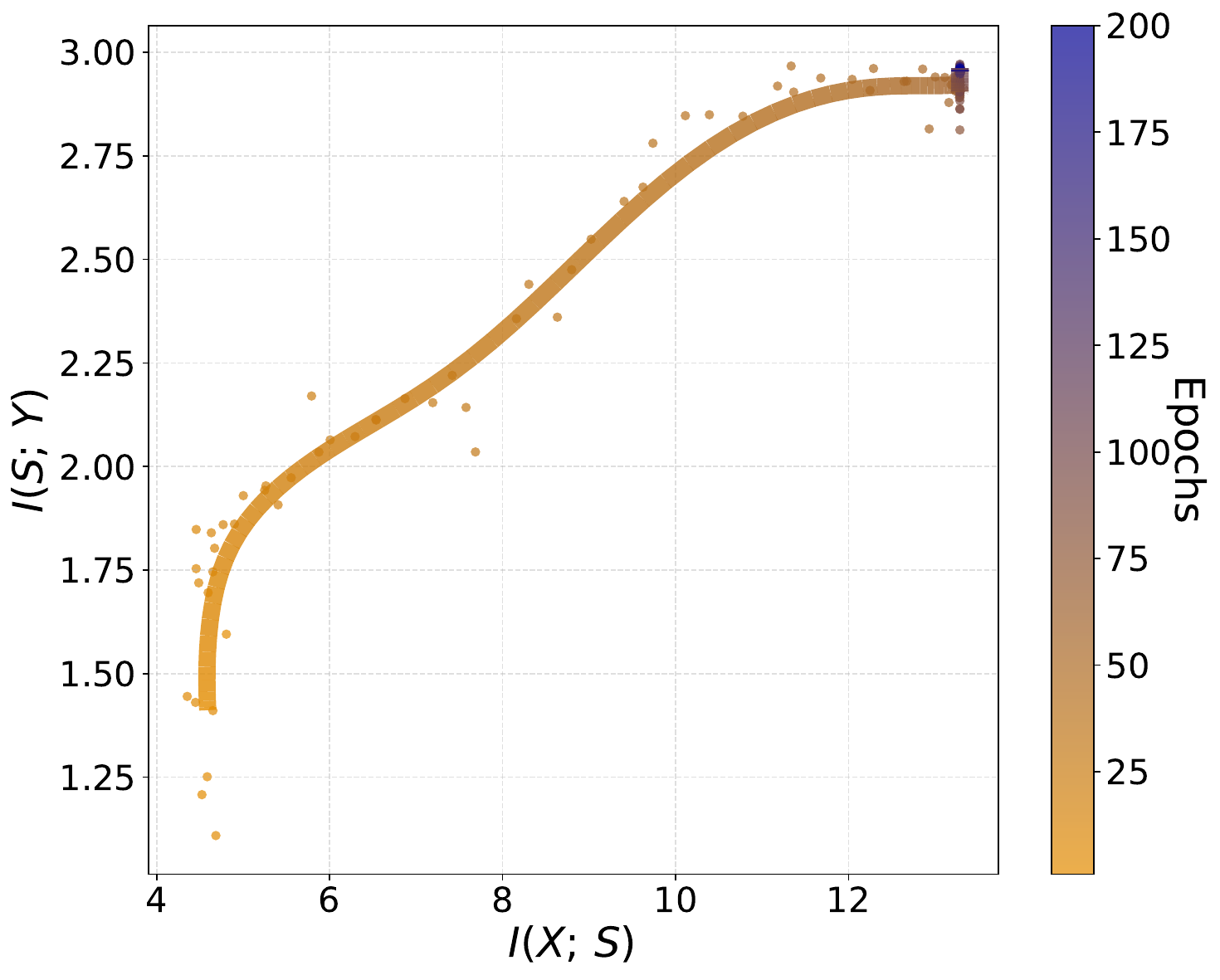}
        \caption{Standard SNN on Fashion-MNIST}
        \label{fig:ip_baseline_fmnist}
    \end{subfigure}
    \hfill
    \begin{subfigure}[b]{0.45\textwidth}
        \centering
        \includegraphics[width=\linewidth]{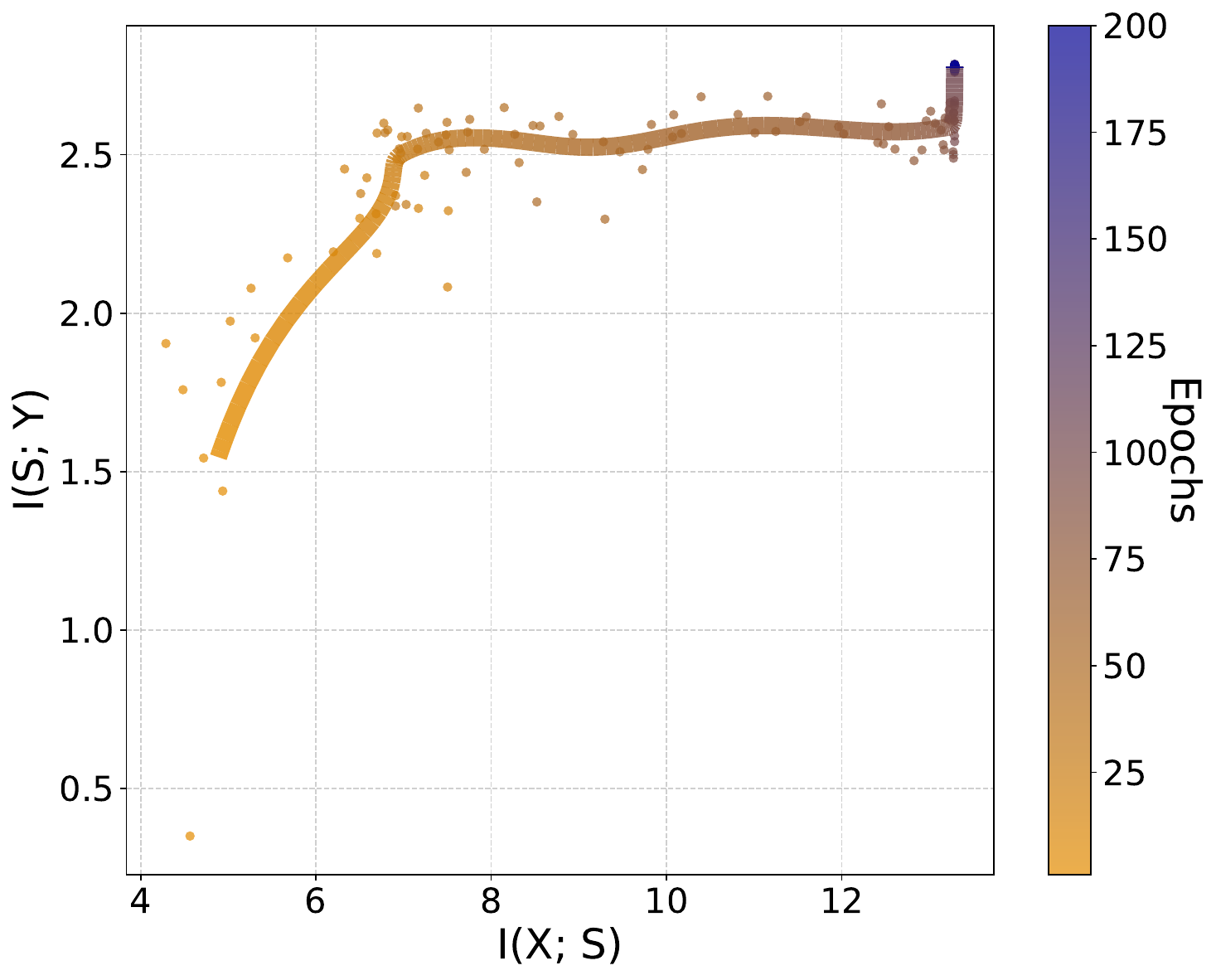}
        \caption{Standard SNN on CIFAR-10}
        \label{fig:ip_baseline_cifar10}
    \end{subfigure}

    \vspace{0.5em}

    \begin{subfigure}[b]{0.45\textwidth}
        \centering
        \includegraphics[width=\linewidth]{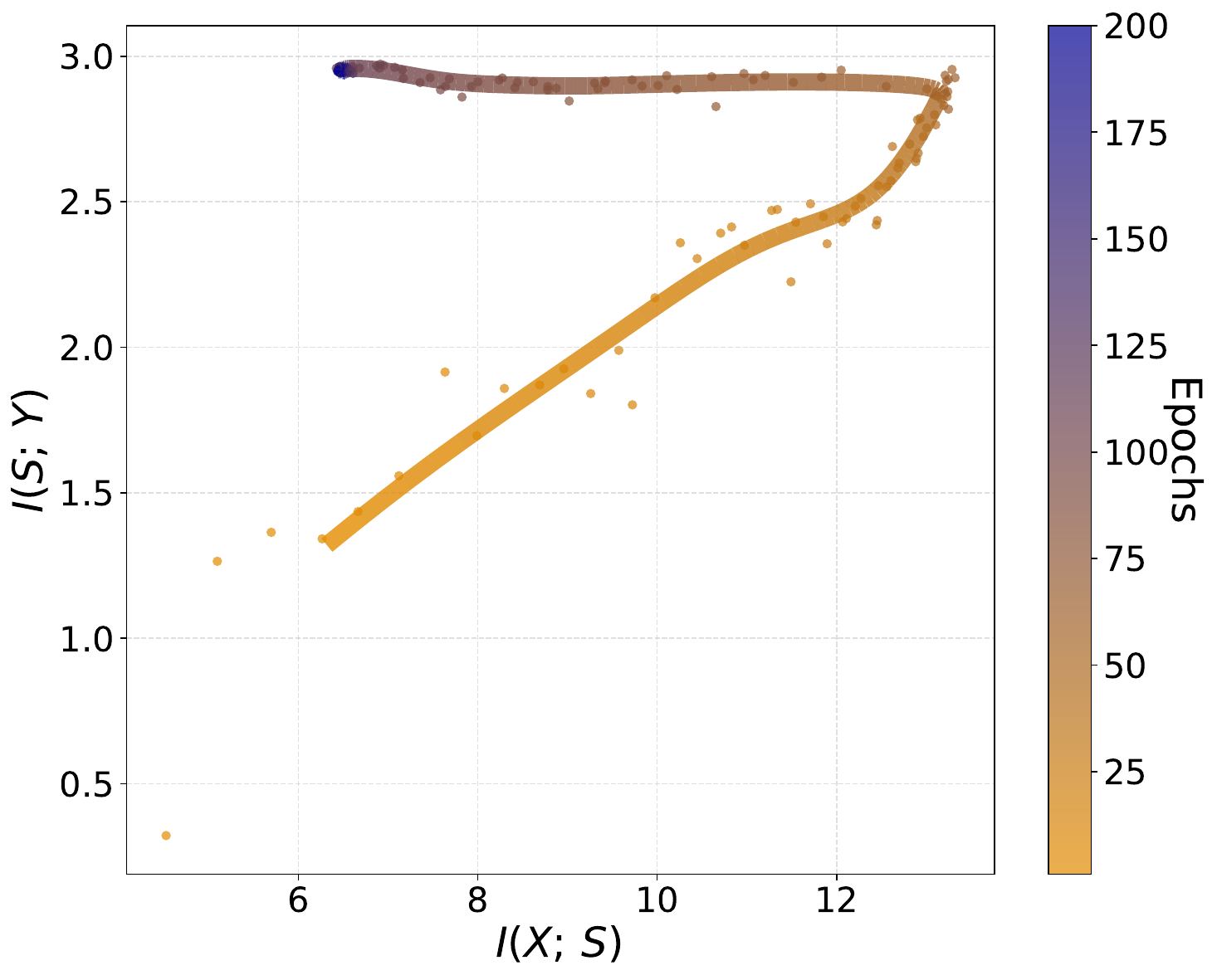}
        \caption{Conventional IB on Fashion-MNIST}
        \label{fig:ip_ib_fmnist}
    \end{subfigure}
    \hfill
    \begin{subfigure}[b]{0.45\textwidth}
        \centering
        \includegraphics[width=\linewidth]{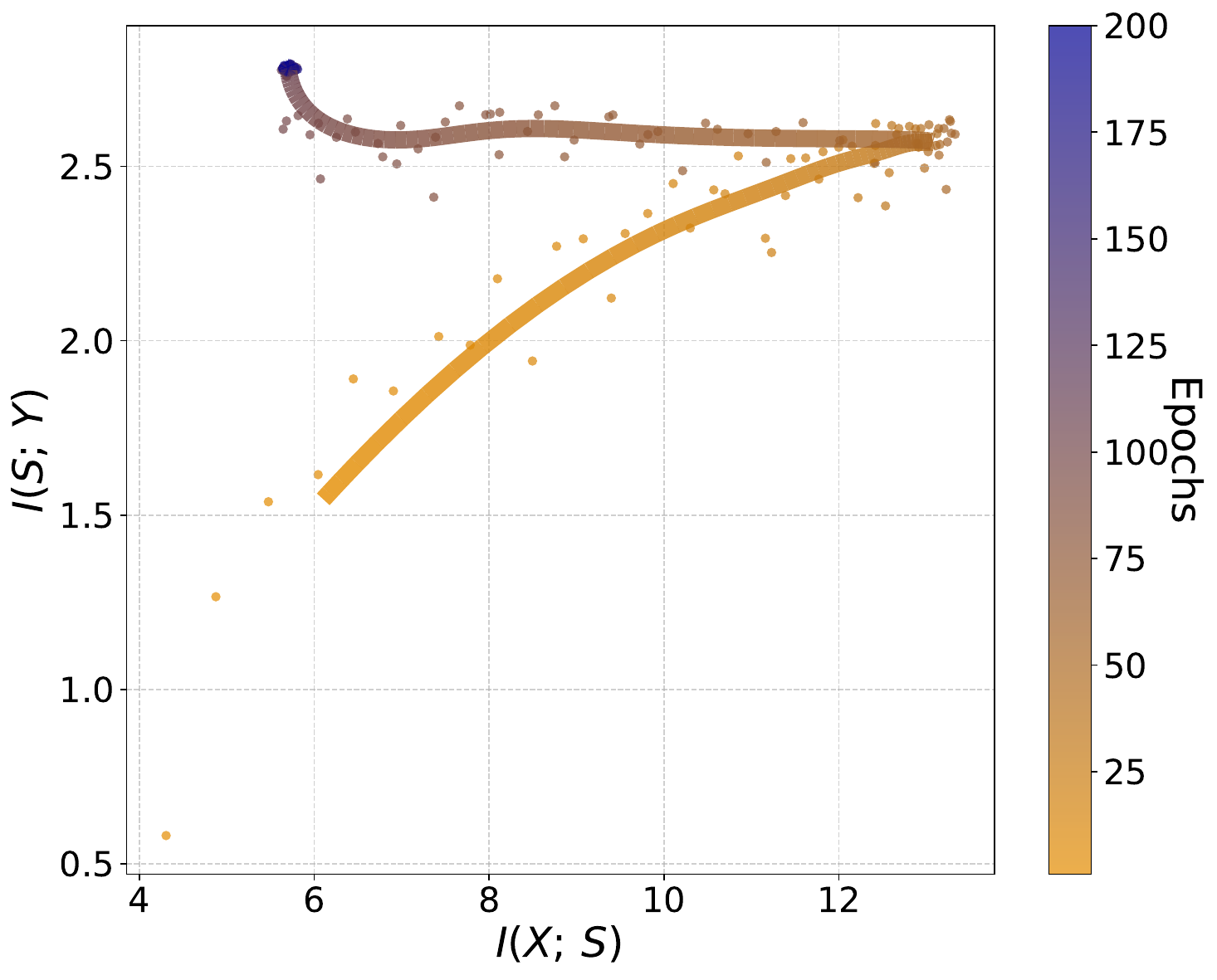}
        \caption{Conventional IB on CIFAR-10}
        \label{fig:ip_ib_cifar10}
    \end{subfigure}
    \caption{Information-plane trajectories of standard SNNs and conventional IB-regularized SNNs on Fashion-MNIST and CIFAR-10. The first row shows the trajectories of standard SNNs, whereas the second row shows those obtained with conventional IB regularization.}
    \label{fig:ip_comparison}
\end{figure*}

Fig.~\ref{fig:ip_comparison} compares the information-plane trajectories of standard SNNs and conventional IB-regularized SNNs on Fashion-MNIST and CIFAR-10. The first row shows the results of the standard SNNs, whereas the second row shows the results obtained after introducing conventional IB regularization.

As shown in the first row of Fig.~\ref{fig:ip_comparison}, the trajectories of the standard SNNs generally move upward and to the right before gradually approaching saturation. During the early stage of training, both $I(X;S)$ and $I(S;Y)$ increase rapidly, indicating that the networks progressively extract input-dependent features that are useful for classification. During the middle and late stages, however, $I(S;Y)$ becomes relatively stable, while $I(X;S)$ remains in a high-information region or continues to increase slowly. Under the adopted estimator and experimental settings, no evident compression phase is observed in the standard SNNs.

This behavior suggests that the learned spike representations may retain substantial input-dependent information even after the label-related information has largely stabilized. Such retained information may include input-specific details, local variations, or perturbation-sensitive components, some of which may not be essential for the final classification decision. Although the information-plane trajectories alone do not establish that all retained information is redundant or harmful, they indicate that conventionally trained SNN representations may benefit from an explicit mechanism that controls unnecessary input dependence.

In contrast, the second row of Fig.~\ref{fig:ip_comparison} shows that the trajectories obtained with conventional IB regularization exhibit a different pattern. During the early stage of training, the trajectories still move upward and to the right, indicating that the networks first acquire label-predictive information. During the middle and late stages, the trajectories gradually bend toward the left: $I(X;S)$ decreases from a relatively high-information region, while $I(S;Y)$ remains at a comparatively high level. This fold-back behavior is consistent with the intended effect of conventional IB regularization, which encourages the representation to reduce its dependence on the input while retaining information relevant to label prediction. 

However, conventional IB achieves compression by directly penalizing the total mutual information $I(X;S)$. As discussed in the following section, under the supervised Markov assumption $Y \leftrightarrow X \leftrightarrow S$ can be decomposed into the label-predictive component $I(S;Y)$ and the label-conditioned component $I(X;S\mid Y)$. Therefore, directly compressing $I(X;S)$ does not explicitly distinguish predictive information from input-specific information that remains after conditioning on the label. When the compression constraint is excessively strong, useful discriminative information may also be suppressed, potentially leading to a robustness--accuracy trade-off. 

These observations motivate the proposed SCIB framework. Instead of indiscriminately penalizing the total dependence between the input and the representation, SCIB uses the conditional mutual information $I(X;S\mid Y)$ as its compression target. In this way, the regularization is directed toward input-specific variations that remain after conditioning on the class label, while the predictive term encourages the retention of label-relevant information.

\section{Methodology}
This section presents the theoretical motivation, variational formulation, learnable class-conditional prior, and optimization procedure of the proposed Spiking Conditional Information Bottleneck (SCIB) framework.
\subsection{Limitations of information bottleneck}

In supervised SNN representation learning, we assume the Markov chain $Y \leftrightarrow X \leftrightarrow S$: the representation $S$ is produced from the input $X$, and labels are used only during training for supervision and regularization. Under this assumption, $S$ and $Y$ are conditionally independent given $X$, i.e., $I(S;Y|X)=0$. Applying the chain rule of mutual information gives
\begin{equation}
I(X;S)=I(S;Y)+I(X;S|Y),
\label{eq:mi_chain_rule}
\end{equation}
where $I(S;Y)$ corresponds to label-relevant predictive information, and $I(X;S|Y)$ measures the residual dependence between $X$ and $S$ after the label $Y$ is given. Thus, $I(X;S|Y)$ measures the input-specific information retained in $S$ after conditioning on the label. This term may include nuisance factors or perturbation-sensitive components, but it may also contain useful intra-class variations such as pose, texture, style, subclass structure, or temporal dynamics.

This decomposition clarifies a limitation of conventional IB regularization. Standard IB compresses representations by penalizing $I(X;S)$, but Eq.~\eqref{eq:mi_chain_rule} shows that $I(X;S)$ contains both predictive information and label-conditioned redundancy. Therefore, directly minimizing $I(X;S)$ does not explicitly distinguish predictive information from other input-dependent factors; when the compression strength is large, it may suppress useful discriminative features together with nuisance variations. Related conditional bottleneck formulations have been used to separate task-relevant or shared information from unnecessary input dependence. The Conditional Entropy Bottleneck formalizes supervised compression through $I(Z;X|Y)$~\cite{Conditional2020Fischer}, while recent sequential models employ conditional mutual-information terms to suppress stream-specific information from shared dynamic representations~\cite{InfoDPCCA2025Tang}. In contrast, SCIB conditions directly on the class label and formulates $I(X;S|Y)$ as label-conditioned redundancy in supervised spike representations.

To formalize the desired representation, we consider the notion of a minimal sufficient representation. Under the Markov chain $Y \leftrightarrow X \leftrightarrow S$, the data processing inequality gives $I(S;Y) \leq I(X;Y)$~\cite{Elements2011Nalewajski}. Following the minimal-sufficiency principle, a desirable representation $S^{*}$ should satisfy two conditions.
\begin{equation}
I(S^{*};Y)=I(X;Y),
\end{equation}
which means that $S^{*}$ preserves all label-relevant information contained in $X$. Second, among all sufficient representations, it should be minimal with respect to the retained input information. Since Eq.~\eqref{eq:mi_chain_rule} gives
\begin{equation}
I(X;S^{*})=I(S^{*};Y)+I(X;S^{*}|Y),
\end{equation}
any sufficient representation satisfies
\begin{equation}
I(X;S^{*})=I(X;Y)+I(X;S^{*}|Y)\geq I(X;Y).
\end{equation}

Therefore, the smallest possible value of $I(X;S^{*})$ under sufficiency is $I(X;Y)$, which is attained when $I(X;S^{*}|Y)=0$. The ideal minimal sufficient representation thus satisfies
\begin{equation}
I(X;S^{*}) = I(S^{*};Y) = I(X;Y),
\label{eq:optimal}
\end{equation}
indicating that a minimal sufficient representation should preserve predictive information while minimizing input-specific information beyond the label.

A second limitation appears in the variational implementation of conventional IB. Standard VIB commonly adopts a label-agnostic unimodal Gaussian prior such as $\mathcal{N}(0,I)$. When applied to class-structured spike representations, this prior pulls posteriors from all classes toward a single global anchor, which may limit representation expressiveness and weaken inter-class separability. These two limitations motivate SCIB to use a conditional compression objective together with a learnable class-conditional prior.

\begin{figure}
    \centering
    \includegraphics[width=\columnwidth]{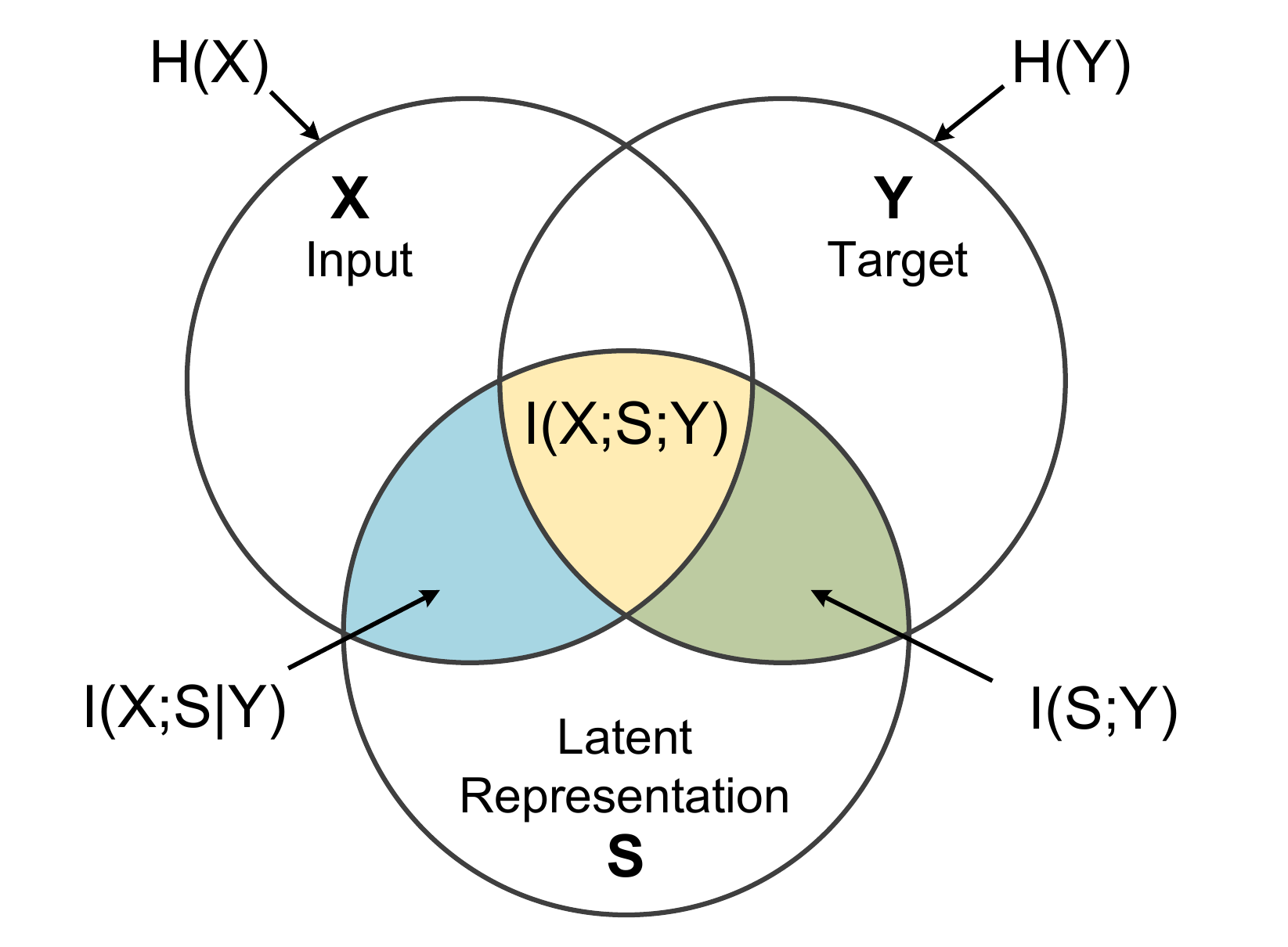}
    \caption{Information diagram for the SCIB framework.}
    \label{fig:InformationDiagram}
\end{figure}

As illustrated in Fig.~\ref{fig:InformationDiagram}, conventional IB minimizes the whole input--representation dependence $I(X;S)$, whereas SCIB aims to compress the residual input information that remains after conditioning on the label. This replaces the indiscriminate compression term $I(X;S)$ with a label-conditioned compression term motivated by $I(X;S|Y)$.

\subsection{SCIB objective function}

Based on the analysis above, reducing \(I(X;S|Y)\) provides a way to regularize input-specific information that remains after conditioning on the class label. Compared with the conventional compression objective $I(X;S)$, this choice has two advantages. First, it separates compression from prediction more explicitly, because $I(X;S|Y)$ corresponds to the part of $I(X;S)$ beyond the predictive information $I(S;Y)$. Second, it provides an interpretable compression target: a smaller  $I(X;S|Y)$ indicates weaker dependence between individual inputs and representations after the label is given.

To apply this principle to spike representations, we use the Leaky Integrate-and-Fire (LIF) neuron as the basic feature extraction unit. At discrete timestep $t$, the membrane potential $V_t$ and the instantaneous output spike $S_t$ follow
\begin{equation}
V_{t} = \tau V_{t-1} + W \cdot S_{t-1},
\label{eq:lif_v}
\end{equation}
\begin{equation}
S_{t} = H(V_{t} - V_{\text{th}}),
\label{eq:lif_s}
\end{equation}
where $\tau$ denotes the membrane-potential decay constant, $W$ is the synaptic weight matrix, $V_{\text{th}}$ is the firing threshold, and $H(\cdot)$ is the Heaviside step function. Over $T$ timesteps, the SNN encoder extracts a high-dimensional latent spike representation from $X$, modeled by the stochastic encoder distribution $p_{\theta}(s|x)$.

We define the SCIB information-theoretic objective to be maximized as
\begin{equation}
\mathcal{J}_{\mathrm{SCIB}} = \max I(S;Y) - \beta \cdot I(X;S|Y),
\label{eq:scib}
\end{equation}
where $\beta$ controls the strength of conditional compression. This objective encourages the representation to preserve label-relevant predictive information while suppressing input-specific redundancy conditioned on the label.

\subsection{Variational bound derivation}

The SCIB information-theoretic objective in Eq.~\eqref{eq:scib} involves $I(X;S|Y)$ and $I(S;Y)$, which are intractable in high-dimensional feature spaces; therefore, the practical training loss optimizes variational bounds rather than the exact information quantities. We derive tractable variational bounds for both terms.

For the conditional compression term, the Markov structure gives $p_{\theta}(s|x,y)=p_{\theta}(s|x)$, and
\begin{equation}
    I(X;S|Y) = \mathbb{E}_{p(x,y)}\!\left[   D_{\mathrm{KL}}\!\left(p_{\theta}(s|x)
    \,\|\, p(s|y)\right)\right].
    \label{eq:cmi_kl}
\end{equation}
Since the true marginal conditional distribution $p(s|y)$ is intractable, we introduce a variational prior $r_{\phi}(s|y)$ to approximate it. Using the non-negativity of KL divergence, we obtain

\begin{equation}
    \begin{aligned}
        I(X;S|Y) 
        &= \mathbb{E}_{p(x,y)}\!\left[D_{\mathrm{KL}}\!\left(
            p_{\theta}(s|x)\,\|\,r_{\phi}(s|y)\right)\right]\\
        &\quad - \mathbb{E}_{p(y)}\!\left[D_{\mathrm{KL}}\!\left(
            p(s|y)\,\|\,r_{\phi}(s|y)\right)\right] \\
        &\leq \mathbb{E}_{p(x,y)}\!\left[D_{\mathrm{KL}}\!\left(
            p_{\theta}(s|x)\,\|\,r_{\phi}(s|y)\right)\right]
    \end{aligned}.
    \label{eq:vub}
\end{equation}

Eq.~\eqref{eq:vub} replaces the intractable marginal conditional distribution $p(s|y)$ with the learnable parameterized distribution $r_{\phi}(s|y)$, yielding a tractable upper bound rather than the exact conditional mutual information. The bound is tight when $r_{\phi}(s|y) = p(s|y)$; therefore, learning a high-quality conditional prior $r_{\phi}(s|y)$ is critical for tightening the upper bound.

Variational lower bound for the prediction term. For the prediction term $I(S;Y)$, we introduce a variational classifier $q_{\psi}(y|s)$ as an approximation to the true predictive distribution $p(y|s)$. Using the relation $I(S;Y) = H(Y) - H(Y|S)$ and the fact that cross-entropy upper-bounds conditional entropy, we derive:
\begin{equation}
    I(S;Y) \geq H(Y) + \mathbb{E}_{p(x,y)}\!\left[
    \log q_{\psi}(y|s)\right].
    \label{eq:vlb}
\end{equation}

Since the empirical entropy $H(Y)$ of the dataset label distribution is constant during training, maximizing the lower bound on predictive information is equivalent to minimizing the standard cross-entropy loss 
$\mathcal{L}_{\mathrm{CE}} = -\mathbb{E}_{p(x,y)}
[\log q_{\psi}(y|s)]$.

Combining the variational upper bound of Eq.~\eqref{eq:vub} with the variational lower bound of Eq.~\eqref{eq:vlb}, maximizing the intractable objective in Eq.~\eqref{eq:scib} is converted into minimizing the following surrogate loss function amenable to end-to-end optimization:
\begin{equation}
    \mathcal{L}_{\mathrm{SCIB}} = \mathcal{L}_{\mathrm{CE}}
    + \beta\,\mathbb{E}_{p(x,y)}\!\left[
    D_{\mathrm{KL}}\!\left(p_{\theta}(s|x)
    \,\|\,r_{\phi}(s|y)\right)\right],
    \label{eq:scib_surrogate}
\end{equation}
where the hyperparameter $\beta > 0$ controls the conditional compression strength in the feature space. The $\mathcal{L}_{\mathrm{CE}}$ term in Eq.~\eqref{eq:scib_surrogate} encourages the representation $S$ to retain predictive information for classification decisions, while the KL divergence term pulls the sample-level encoder distribution $p_{\theta}(s|x)$ towards the class-conditional prior $r_{\phi}(s|y)$ of the corresponding class, encouraging class-conditioned compactness. In this way, the representation is encouraged to reduce its dependence on input-specific variations after conditioning on the label, which may help reduce perturbation-sensitive components without explicitly assuming a particular noise type.

\subsection{Learnable class-conditional prior}
\label{sec:class_conditional_prior}
The quality of the variational upper bound depends on how well $r_{\phi}(s|y)$ approximates $p(s|y)$. Conventional variational inference typically adopts a label-agnostic standard Gaussian as the prior, but this choice may encourage feature posteriors of different classes to share the same global anchor, potentially weakening inter-class separability in the representation space and increasing the risk of representation degeneration and inter-class confusion. To address this, we propose a Learnable Class-Conditional Prior that assigns each class an independent distributional anchor in the latent space.

For a classification task with $K$ classes, we assign each class $k \in \{1, \ldots, K\}$ a set of independent learnable parameters: a mean vector $\mu_{k} \in \mathbb{R}^{d}$ and a variance vector $\sigma_{k}^{2} \in \mathbb{R}_{+}^{d}$, where $d$ is the latent feature dimensionality. The variational prior is defined as a set of class-specific diagonal Gaussian distributions:
\begin{equation}
    r_{\phi}(s \mid y=k) = \mathcal{N}\!\left(
    s;\, \mu_{k},\, \mathrm{diag}(\sigma_{k}^{2})\right).
    \label{eq:prior}
\end{equation}

After marginalizing over labels, the class-conditional prior can be interpreted as a Gaussian mixture:
\begin{equation}
    r_{\phi}(s) = \sum_{k=1}^{K} p(y=k)\,
    r_{\phi}(s \mid y=k).
    \label{eq:mixture_prior}
\end{equation}
The SNN encoder $p_{\theta}(s|x)$ extracts sample-level latent feature mean $\mu_{x}$ and variance $\sigma_{x}^{2}$ from the given input $X$ through the spatiotemporal integration dynamics of multiple spiking layers, and constructs the variational posterior via the reparameterization trick:
\begin{equation}
    p_{\theta}(s \mid x) = \mathcal{N}\!\left(
    s;\, \mu_{x},\, \mathrm{diag}(\sigma_{x}^{2})\right).
    \label{eq:posterior}
\end{equation}

Since both the variational prior and the variational posterior are parameterized as diagonal Gaussians, the KL divergence regularization term in Eq.~\eqref{eq:scib_surrogate} admits an efficient closed-form analytical solution, bypassing high-variance Monte Carlo sampling. For a spike representation of feature dimensionality $d$, given input $x$ and its corresponding label $y = k$, the analytical form of the KL divergence is:

\begin{equation}
    \begin{aligned}
        & D_{\mathrm{KL}}\!\left[p_{\theta}(s \mid x)
        \,\|\, r_{\phi}(s \mid y=k)\right] \\
        ={} & \frac{1}{2}\sum_{j=1}^{d}\left(
          \log\frac{\sigma_{k,j}^{2}}{\sigma_{x,j}^{2}}
          + \frac{\sigma_{x,j}^{2} + 
          (\mu_{x,j} - \mu_{k,j})^{2}}{\sigma_{k,j}^{2}}
          - 1\right)
    \end{aligned}.
    \label{eq:kl_closed}
\end{equation}

Eq.~\eqref{eq:kl_closed} enters directly as the regularization component of the loss function, participating in the joint gradient updates of the SNN synaptic weights $\theta$, classifier parameters $\psi$, and prior parameters $\phi = \{\mu_{k},\, \mathrm{diag}(\sigma_{k}^{2})\}_{k=1}^{K}$.

From a representation-learning perspective, this regularization has two effects. First, it encourages intra-class compactness by pulling the posterior mean $\mu_x$ toward the prior center $\mu_k$ of the corresponding class and constraining posterior dispersion through the variance terms. Second, when combined with the cross-entropy objective, it may help maintain inter-class separability because different classes are assigned independent learnable distributional anchors, reducing the tendency of classes to be pulled toward a single standard Gaussian prior.

\subsection{Overall framework and algorithm}
\begin{figure*}[t]
    \centering
    \includegraphics[width=\textwidth]{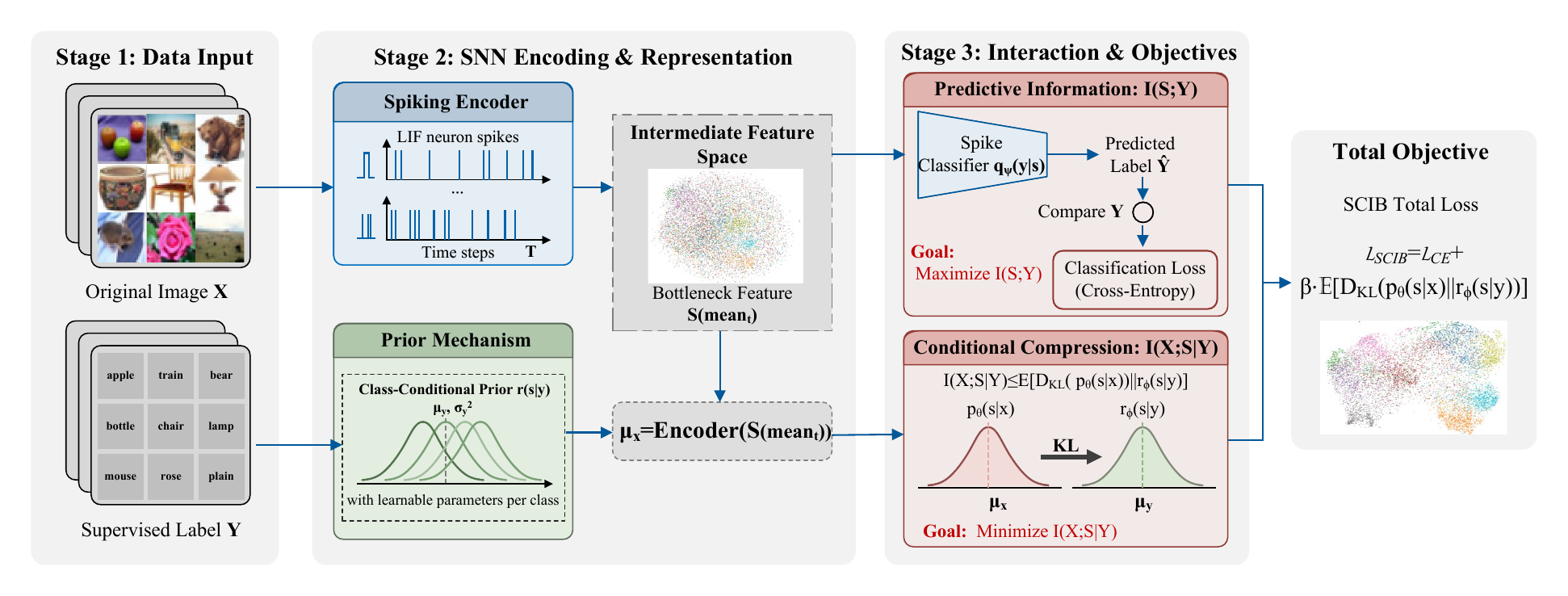}
    \caption{The overall architecture of the SCIB framework.}
    \label{fig:scib_framework}
\end{figure*}

Based on the theoretical derivations and mechanism designs presented above, this section presents the overall architecture and complete optimization algorithm of the SCIB framework. Fig.~\ref{fig:scib_framework} illustrates the end-to-end training pipeline of SCIB.

First, the input $X$ is processed by the SNN encoder over $T$ timesteps to obtain the latent spike representation and the posterior statistics $(\mu_x,\sigma_x^2)$. Second, the ground-truth label $y$ is used only during training to select the corresponding class-conditional prior $r_{\phi}(s|y)$, while the sampled latent variable is passed to the classifier for prediction. Finally, the cross-entropy loss and the closed-form KL regularizer are combined according to Eq.~\eqref{eq:scib_surrogate}, and the SNN encoder parameters $\theta$, classifier parameters $\psi$, and prior parameters $\phi$ are jointly optimized via BPTT. During inference, the class-conditional prior is not used, and prediction follows the standard feedforward SNN pathway. The complete optimization procedure is summarized in Algorithm~\ref{alg:scib}.

\begin{algorithm}
\caption{Optimization of SCIB}
\label{alg:scib}
\SetAlgoLined
\SetKwInOut{Input}{Input}
\SetKwInOut{Output}{Output}

\Input{Training dataset $\mathcal{D} = \{(x_i, y_i)\}_{i=1}^{N}$; 
     number of classes $K$; timestep $T$; parameters $\theta$ (including decay constant $\tau$, 
    threshold $V_{\mathrm{th}}$, weights $W$); classifier parameters $\psi$; prior parameters  $\phi = \{\mu_k, \sigma_k^2\}_{k=1}^{K}$; compression coefficient $\beta$; learning rate $\eta$; training epoch $E$; mini-batch $B$}
\Output{Optimized parameters $\theta^{*}$, classifier parameters $\psi^{*}$, prior parameters $\phi^{*}$}

\BlankLine
Initialize SNN weights $\theta$ and classifier parameters $\psi$\;

\For{$k \leftarrow 1$ \KwTo $K$}{
    Initialize prior mean $\mu_k \sim \mathcal{N}(0,\, 0.01 I)$ and 
    log-variance $\log \sigma_k^2 \sim \mathcal{N}(0,\, 0.01 I)$\;
}

\For{$\mathrm{epoch} \leftarrow 1$ \KwTo $E$}{
    \For{each mini-batch $\{(x_i, y_i)\}_{i=1}^{B} \subseteq \mathcal{D}$}{
        Initialize membrane potentials $V_l^{(0)} = 0$ and spikes $S_l^{(0)} = 0$ for all layers $l$\;
        
        \For{$t \leftarrow 1$ \KwTo $T$}{
            \For{$l \leftarrow 1$ \KwTo $L$}{
                LIF dynamics: $V_l^{(t)} = \tau V_l^{(t-1)} \odot (1 - S_l^{(t-1)}) + W_{l-1,l}\, S_{l-1}^{(t)}$\;
                Spike emission: $S_l^{(t)} = H(V_l^{(t)} - V_{\mathrm{th}})$\;
            }
        }
        
        Aggregate spikes over time to obtain latent representation $\bar{S}$\;
        Extract posterior statistics: $(\mu_x,\, \log\sigma_x^2) = \mathrm{Encoder}(\bar{S};\, \theta)$\;
        Reparameterize: $z = \mu_x + \sigma_x \odot \varepsilon$, \quad $\varepsilon \sim \mathcal{N}(0, I)$\;
        Compute prediction: $\hat{y} = \mathrm{Classifier}(z;\, \psi)$\;
        Compute cross-entropy loss: $\mathcal{L}_{\mathrm{CE}} = -\log q_{\psi}(y \mid z)$\;
        Index class-conditional prior via label $y$: $r_{\phi}(z \mid y) = \mathcal{N}(\mu_y,\, \mathrm{diag}(\sigma_y^2))$\;
        Compute conditional compression regularizer: $\mathcal{R}_{\mathrm{KL}} = D_{\mathrm{KL}}\!\left(p_{\theta}(s|x) \,\|\, r_{\phi}(s|y)\right)$\;
        Compute SCIB loss: $\mathcal{L}_{\mathrm{SCIB}} = \mathcal{L}_{\mathrm{CE}} + \beta \cdot \mathcal{R}_{\mathrm{KL}}$\;
        Compute gradients via BPTT: $\nabla_{\theta}\mathcal{L}_{\mathrm{SCIB}}$, $\nabla_{\psi}\mathcal{L}_{\mathrm{SCIB}}$, $\nabla_{\phi}\mathcal{L}_{\mathrm{SCIB}}$\;
        Update parameters: \quad $\theta \leftarrow \theta - \eta\, \nabla_{\theta}\mathcal{L}_{\mathrm{SCIB}}$, $\psi \leftarrow \psi - \eta\, \nabla_{\psi}\mathcal{L}_{\mathrm{SCIB}}$, $\phi \leftarrow \phi - \eta\, \nabla_{\phi}\mathcal{L}_{\mathrm{SCIB}}$\;
    }
}
\end{algorithm}

\section{Experiments and results}

\subsection{Experimental setup}

Environment and training configuration. All experiments are implemented using PyTorch 1.12 and Python 3.9 on Ubuntu 20.04, with training and evaluation conducted on an NVIDIA A100 GPU. All models are trained with the SGD optimizer with momentum 0.9, an initial learning rate of 0.1, and a cosine annealing schedule for learning rate decay. The batch size is 128. The compression strength hyperparameter $\beta$ of SCIB is determined via validation set sensitivity analysis, with a default value of $\beta = 0.4$; for hardware noise experiments, $\beta = 0.2$ is adopted to account for the differential sensitivity of membrane potential perturbations to the compression constraint.

\textbf{Datasets.} To evaluate SCIB across varying levels of data complexity, three widely used image classification benchmark datasets are selected: Fashion-MNIST~\cite{Fashion2017Xiao}, CIFAR-10~\cite{Learning2009Krizhevsky}, and CIFAR-100~\cite{Learning2009Krizhevsky}. The three datasets exhibit an increasing relationship in terms of class granularity and feature space dimensionality.

\textbf{Network architecture.} The backbone network is a VGG-9 SNN architecture adapted for the spike domain, comprising 7 convolutional layers (Conv + BN + LIF) and 1 fully connected layer, with the number of timesteps set to $T = 4$; all spiking neurons adopt the LIF model. In adversarial attack experiments, comparisons are conducted under two timestep settings, $T = 4$ and $T = 8$, to evaluate the influence of temporal depth and align with the configuration of FEEL-SNN.

\textbf{Baseline methods.} Since random perturbations and adversarial attacks correspond to different threat models, two groups of comparison baselines are employed. For \textit{noise robustness experiments}, we compare SCIB with a standard SNN baseline and several IB-based SNN baselines:
(1) Without IB (w.o. IB)~\cite{Surrogate2019Neftci}: a standard SNN trained with the surrogate gradient method using cross-entropy loss and no information compression constraint, serving as the ablation control;
(2) IB~\cite{Deep2016Alemi}: the conventional information bottleneck applied directly to SNN training, with a standard Gaussian $\mathcal{N}(0,I)$ as the variational prior to minimize the variational upper bound of $I(X;S)$;
(3) NBP-SNN~\cite{Efficient2021Guo}: a non-backpropagation SNN training method based on the IB principle, computing local losses using the membrane potential of each layer as the information representation;
(4) SIB and QIB~\cite{SNIB2023Yang}: the squared and quartic variants of the nonlinear information bottleneck within the SNIB framework. For \textit{adversarial attack experiments}, we compare SCIB with two reference methods:
(1) Vanilla SNN~\cite{Surrogate2019Neftci}: standard cross-entropy training with no defense mechanism;
(2) FEEL-SNN~\cite{FEEL2024Xu}: a frequency-encoding-based robust SNN method with an evolutionary leak factor, where the frequency parameter is set to $r = [16, 14, 12, 10]$ for $T = 4$ and $r = [16, 14, 12, 10, 8, 6, 4, 2]$ for $T = 8$. SCIB does not employ any adversarial training mechanism; its adversarial robustness is evaluated as a by-product of representation-level regularization rather than as a replacement for dedicated adversarial training.

\textbf{External environmental noise}. Three representative types of external environmental noise are introduced to evaluate the interference resistance of SCIB in real physical environments. The Normalized Noise Intensity (NNI) is defined independently for each noise type.

\begin{enumerate}[(1)]
\item Gaussian noise simulates sensor thermal noise and electronic circuit floor noise. A perturbation drawn from a normal distribution is added to each pixel $x$ in the image:
\begin{equation}
    \bar{x} = x + \mathcal{N}(0, \sigma^{2}),
    \label{eq:gaussian_noise}
\end{equation}
where NNI is defined as the standard deviation $\sigma$, with tested values $NNI=0.25$ and $NNI=1.0$.

\item Salt-and-pepper noise simulates impulse interference caused by dead pixels in event camera sensors or spike transmission errors. A proportion $P$ of pixels in the image are randomly replaced by extreme values, 0 or 1:
\begin{equation}
    \bar{x} = \mathrm{SP}(x, P),
    \label{eq:sp_noise}
\end{equation}
where NNI is defined as the noise density $P$. Two pixel corruption ratios are tested: 10\% (NNI $= 0.1$) and 30\% (NNI $= 0.3$).

\item Rayleigh noise is used to evaluate model robustness under non-Gaussian and asymmetric noise distributions:
\begin{equation}
    \bar{x} = x + \sigma_{\mathrm{ray}}
    \sqrt{-2\ln(1 - u)}, \quad u \sim \mathcal{U}(0, 1),
    \label{eq:rayleigh_noise}
\end{equation}
where $\sigma_{\mathrm{ray}}$ is the scale parameter of the Rayleigh distribution, and NNI is defined as $\sigma_{\mathrm{ray}}$. Two perturbation levels are tested: NNI $= 0.25$ and NNI $= 1.0$.
\end{enumerate}

\textbf{Neuromorphic hardware noise.} When deploying SNNs on edge computing devices or mixed-signal neuromorphic chips, the non-ideal characteristics of physical devices introduce unavoidable internal noise. Two representative hardware-level perturbations are considered.

\begin{enumerate}[(1)]
\item Synaptic weight noise simulates thermal noise and conductance drift occurring during weight updates or state readout in physical synapses. Zero-mean Gaussian perturbations are injected into synaptic weight updates during training:
\begin{equation}
    w_{\mathrm{upd}} = \Delta w + 
    \mathcal{N}(0, \sigma_{\mathrm{SN}}^{2}),
    \label{eq:syn_noise}
\end{equation}
where $\Delta w$ is the ideal update magnitude computed by backpropagation, and $w_{\mathrm{upd}}$ is the weight actually applied to the network after perturbation.

\item Membrane potential noise simulates background perturbations of neuronal membrane potentials caused by crosstalk, leakage currents, or thermal fluctuations in neuromorphic circuits:
\begin{equation}
    \tilde{V} = V + \mathcal{N}(0, \sigma_{\mathrm{MN}}^{2}),
    \label{eq:mem_noise}
\end{equation}
where $V$ is the theoretical membrane potential determined by the LIF neuron dynamics, and $\tilde{V}$ is the perturbed membrane potential. Two NNI levels are tested for both hardware noise types: NNI $= 1$ and NNI $= 5$.
\end{enumerate}

\textbf{Adversarial attack.} Three gradient-based adversarial attack methods are employed to evaluate the adversarial robustness of SCIB. All attacks are conducted under the $L_{\infty}$ norm threat model with perturbation budget $\epsilon \in \{4/255,\; 8/255\}$. The non-differentiability of spiking neuron activation functions is circumvented via surrogate gradient techniques, with gradients approximated by BPTT.

\begin{enumerate}[(1)]
\item Fast Gradient Sign Method (FGSM)~\cite{Explaining2014Goodfellow} is a single-step attack that perturbs the input along the sign of the loss gradient:
\begin{equation}
    \hat{x} = x + \epsilon \cdot 
    \mathrm{sign}\!\left(\nabla_{x}
    \mathcal{L}(f(x), y)\right),
    \label{eq:fgsm}
\end{equation}
where $\mathcal{L}$ is the classification loss, $f$ denotes the SNN model, and $y$ is the ground-truth label.

\item Projected Gradient Descent (PGD)~\cite{Towards2018Madry} is an iterative extension of FGSM that starts from a random initialization within the $\epsilon$-ball and performs multi-step projected gradient updates:
\begin{equation}
    \hat{x}^{0} = x + \mathcal{U}(-\epsilon, +\epsilon),
    \label{eq:pgd_init}
\end{equation}
\begin{equation}
    \hat{x}^{k+1} = \Pi_{\epsilon}\!\left\{
    \hat{x}^{k} + \alpha \cdot \mathrm{sign}\!\left(
    \nabla_{\hat{x}^{k}}\mathcal{L}
    (f(\hat{x}^{k}), y)\right)\right\},
    \label{eq:pgd_update}
\end{equation}
where $\alpha$ is the step size and $\Pi_{\epsilon}$ denotes the projection onto the $L_{\infty}$ $\epsilon$-ball centered at $x$. We set $\alpha = 0.01$ with $k = 7$ iterations, with adversarial examples clipped to both the valid pixel range and the $L_{\infty}$ ball.

\item Basic Iterative Method (BIM)~\cite{Adversarial2017Kurakin} is similar to PGD but starts from the original clean image rather than a random initialization:
\begin{equation}
    \hat{x}^{0} = x,
    \label{eq:bim_init}
\end{equation}
\begin{equation}
    \hat{x}^{k+1} = \mathrm{Clip}_{x,\epsilon}\!\left\{
    \hat{x}^{k} + \alpha \cdot \mathrm{sign}\!\left(
    \nabla_{\hat{x}^{k}}\mathcal{L}
    (f(\hat{x}^{k}), y)\right)\right\},
    \label{eq:bim_update}
\end{equation}
where $\mathrm{Clip}_{x,\epsilon}$ performs pixel-wise clipping to ensure the result always lies within the $L_{\infty}$ $\epsilon$-neighborhood of the original image. The step size and number of iterations are the same as PGD.
\end{enumerate}

\textbf{Evaluation metric.} Classification accuracy is used as the primary evaluation metric across all experiments. The energy consumption analysis additionally employs a theoretical energy model based on Synaptic Operations (SynOps).

\subsection{External environmental noise robustness}

To evaluate the interference resistance of SCIB under real-world physical environmental noise, we conduct systematic external noise robustness tests on Fashion-MNIST~\cite{Fashion2017Xiao}, CIFAR-10~\cite{Learning2009Krizhevsky}, and CIFAR-100~\cite{Learning2009Krizhevsky}, covering three representative noise distributions: Salt-and-Pepper, Gaussian, and Rayleigh. Each noise type is tested at two intensity levels. The classification accuracy of all methods across 18 noise configurations is reported in Table~\ref{tab:noise}.

\begin{table*}[t]
\centering
\caption{Classification accuracy (\%) of all methods under 18 external environmental noise configurations. Boldface and underlining denote the best and second-best results in each row, respectively.}
\label{tab:noise}
\begin{tabular}{cccccccccc}
\hline
\multirow{2}{*}{Noise Type} & 
\multirow{2}{*}{Noise Intensity} & 
\multirow{2}{*}{Dataset} & 
\multicolumn{6}{c}{Accuracy (\%)} \\ 
\cline{4-9}
& & & w.o. IB & IB & NBP-SNN & SIB & QIB & SCIB \\ 
\hline
\multirow{6}{*}{Salt-and-Pepper} 
& \multirow{3}{*}{NNI=0.1} 
    & Fashion-MNIST & 92.96 & \underline{93.09} & 92.85 & 92.94 & 93.08 & \textbf{93.28} \\
&   & CIFAR-10      & 88.38 & 89.07 & 88.84 & 88.75 & \underline{89.33} & \textbf{89.99} \\
&   & CIFAR-100     & 61.52 & 62.32 & 62.84 & \textbf{63.27} & \underline{62.92} & 62.91 \\ 
\cline{2-9}
& \multirow{3}{*}{NNI=0.3} 
    & Fashion-MNIST & 92.39 & 92.50 & \underline{92.65} & 92.47 & 92.48 & \textbf{92.73} \\
&   & CIFAR-10      & 85.59 & 86.61 & 86.01 & \underline{86.88} & 86.31 & \textbf{87.13} \\
&   & CIFAR-100     & 55.90 & 57.08 & 56.92 & \underline{57.49} & 56.61 & \textbf{58.40} \\ 
\hline
\multirow{6}{*}{Gaussian} 
& \multirow{3}{*}{NNI=0.25} 
    & Fashion-MNIST & 91.26 & 92.67 & 91.71 & \underline{92.69} & 92.68 & \textbf{92.77} \\
&   & CIFAR-10      & 88.23 & \textbf{88.52} & 87.52 & 88.38 & 88.40 & \underline{88.44} \\
&   & CIFAR-100     & 61.19 & 62.28 & 61.89 & 61.79 & \underline{62.51} & \textbf{62.89} \\ 
\cline{2-9}
& \multirow{3}{*}{NNI=1.0} 
    & Fashion-MNIST & 87.98 & 88.13 & 87.77 & \underline{89.34} & 89.28 & \textbf{89.86} \\
&   & CIFAR-10      & 78.48 & \underline{81.44} & 80.38 & 81.21 & 81.36 & \textbf{81.78} \\
&   & CIFAR-100     & 50.21 & 50.72 & 50.23 & 50.65 & \underline{51.19} & \textbf{51.32} \\ 
\hline
\multirow{6}{*}{Rayleigh} 
& \multirow{3}{*}{NNI=0.25} 
    & Fashion-MNIST & 88.69 & 88.87 & 88.83 & \underline{89.20} & 89.05 & \textbf{89.51} \\
&   & CIFAR-10      & 83.72 & 84.60 & 84.69 & 85.05 & \underline{85.16} & \textbf{85.62} \\
&   & CIFAR-100     & 54.31 & 55.35 & 55.45 & 55.20 & \underline{56.10} & \textbf{56.29} \\ 
\cline{2-9}
& \multirow{3}{*}{NNI=1.0} 
    & Fashion-MNIST & 84.07 & \textbf{85.07} & 83.97 & 84.75 & 84.53 & \underline{84.99} \\
&   & CIFAR-10      & 64.89 & 65.33 & 64.78 & \underline{66.17} & 64.62 & \textbf{66.52} \\
&   & CIFAR-100     & 45.91 & 45.39 & 45.80 & 45.99 & \underline{46.06} & \textbf{46.60} \\ 
\hline
\end{tabular}
\end{table*}

Salt-and-pepper noise. Under mild corruption (NNI $= 0.1$), the performance of all IB-based methods is comparable. On CIFAR-100, SCIB achieves 62.91\%, slightly below SIB's 63.27\% and QIB's 62.92\%. When the corruption rate increases to the extreme level (NNI $= 0.3$), the performance gap widens: SCIB achieves the best results on all three datasets, reaching 58.40\% on CIFAR-100 and surpassing the second-best SIB by 0.91 percentage points.

Gaussian noise. Under moderate perturbation (NNI $= 0.25$), SCIB performs comparably to all IB baselines, achieving 88.44\% on CIFAR-10 versus IB's 88.52\%. Under stronger perturbation (NNI $= 1.0$), SCIB achieves the best results on all three datasets, leading the unconstrained baseline by approximately 3.3 percentage points on CIFAR-10 and 1.1 percentage points on CIFAR-100.

Rayleigh noise. SCIB achieves the best results across all four configurations on CIFAR-10 and CIFAR-100. Under extreme perturbation on CIFAR-100 (NNI $= 1.0$), IB drops 0.52 percentage points below the unconstrained baseline, whereas SCIB maintains the best accuracy at 46.60\%. This contrast suggests that the class-conditional prior used by SCIB is more suitable than a label-agnostic standard Gaussian prior under asymmetric non-Gaussian perturbations.

\subsection{Neuromorphic hardware noise robustness}

\begin{figure*}[t]
    \centering
    \begin{subfigure}[b]{0.32\textwidth}
        \centering
        \includegraphics[width=\linewidth]{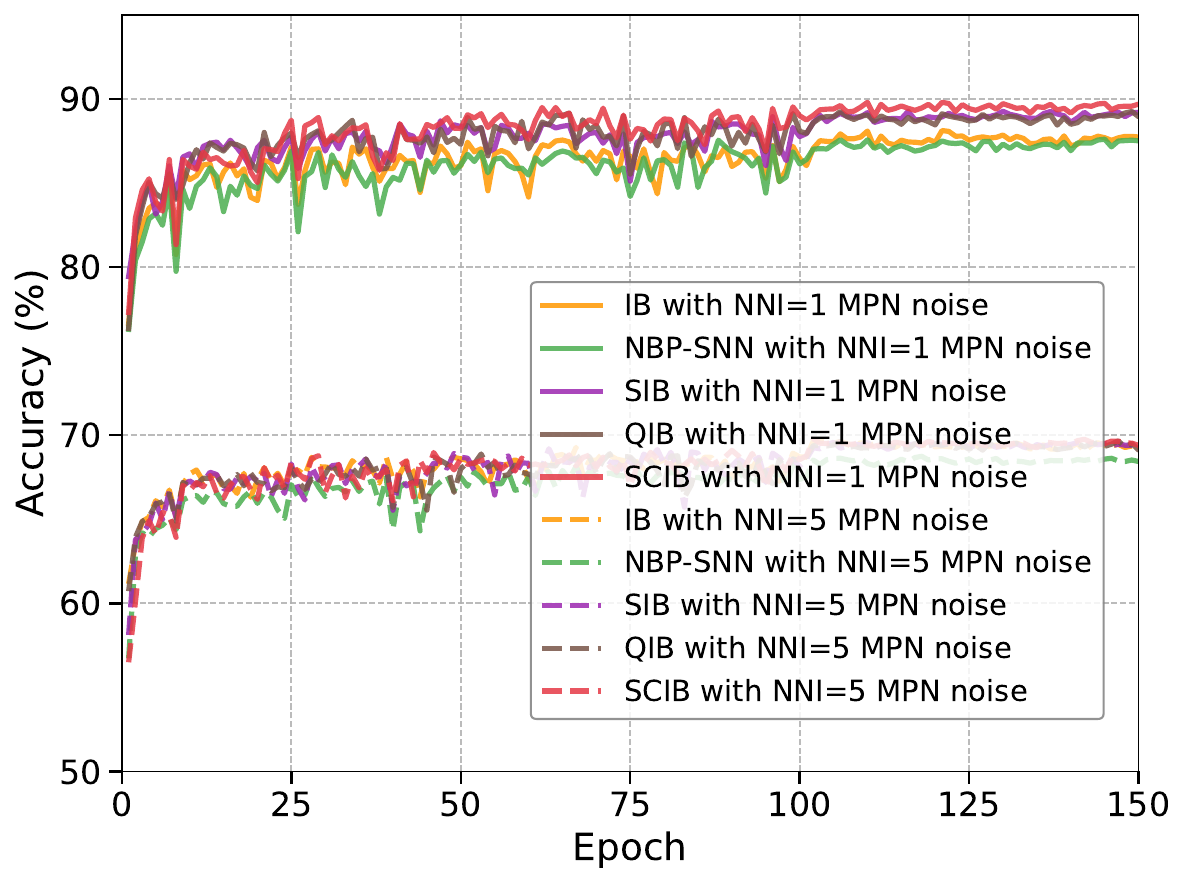}
        \caption{Fashion-MNIST}
        \label{fig:mpn_fashion}
    \end{subfigure}
    \hfill
    \begin{subfigure}[b]{0.32\textwidth}
        \centering
        \includegraphics[width=\linewidth]{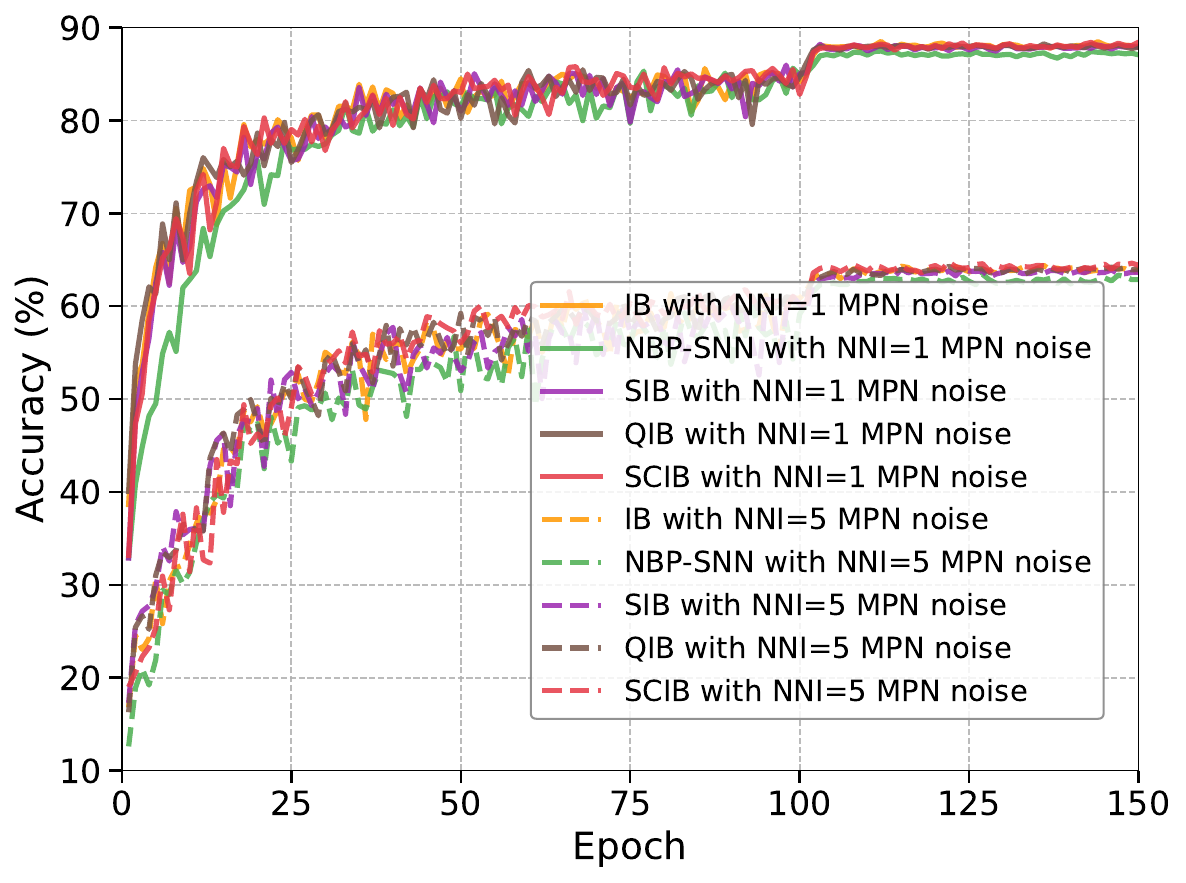}
        \caption{CIFAR-10}
        \label{fig:mpn_cifar10}
    \end{subfigure}
    \hfill
    \begin{subfigure}[b]{0.32\textwidth}
        \centering
        \includegraphics[width=\linewidth]{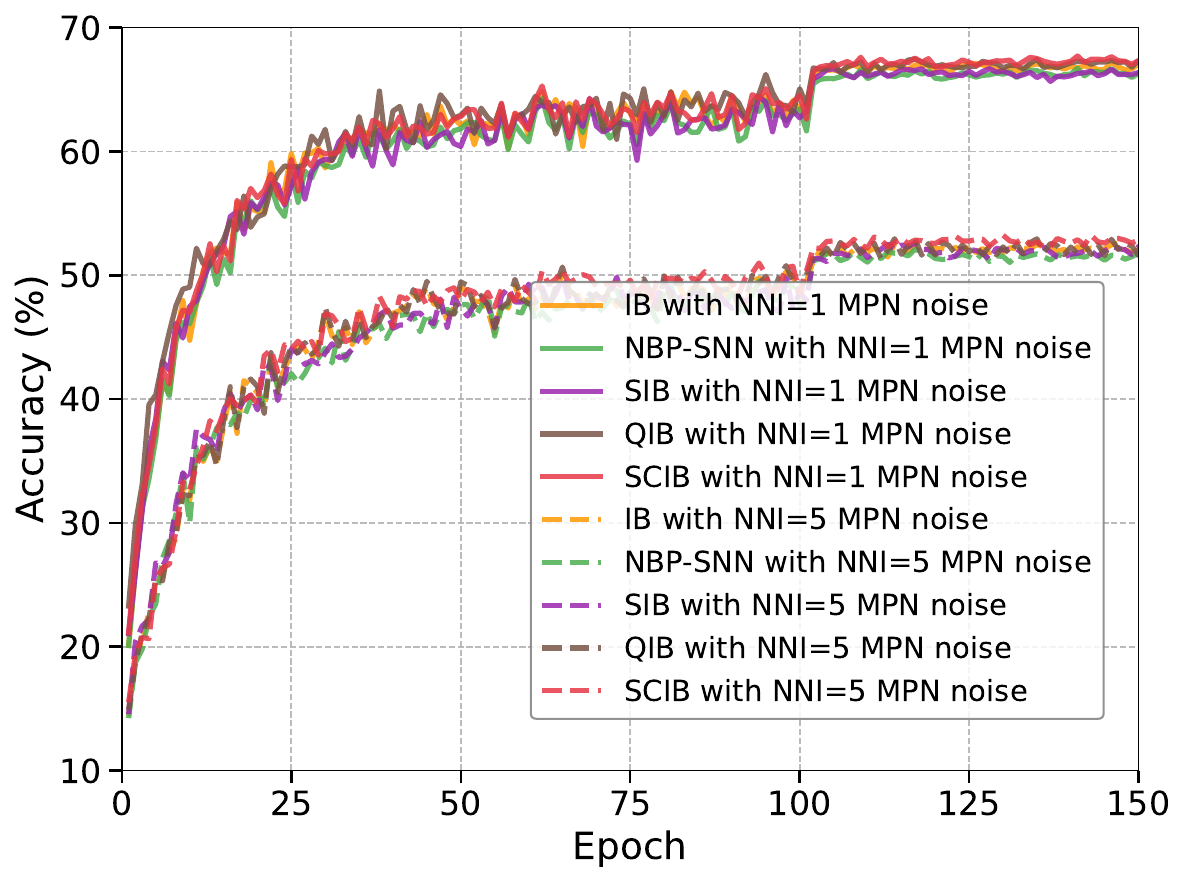}
        \caption{CIFAR-100}
        \label{fig:mpn_cifar100}
    \end{subfigure}
    \caption{Training convergence curves of all methods under membrane potential noise (MPN). Solid and dashed lines correspond to NNI $= 1$ and NNI $= 5$, respectively.}
    \label{fig:mpn}
\end{figure*}

\begin{figure*}[t]
    \centering
    \begin{subfigure}[b]{0.32\textwidth}
        \centering
        \includegraphics[width=\linewidth]{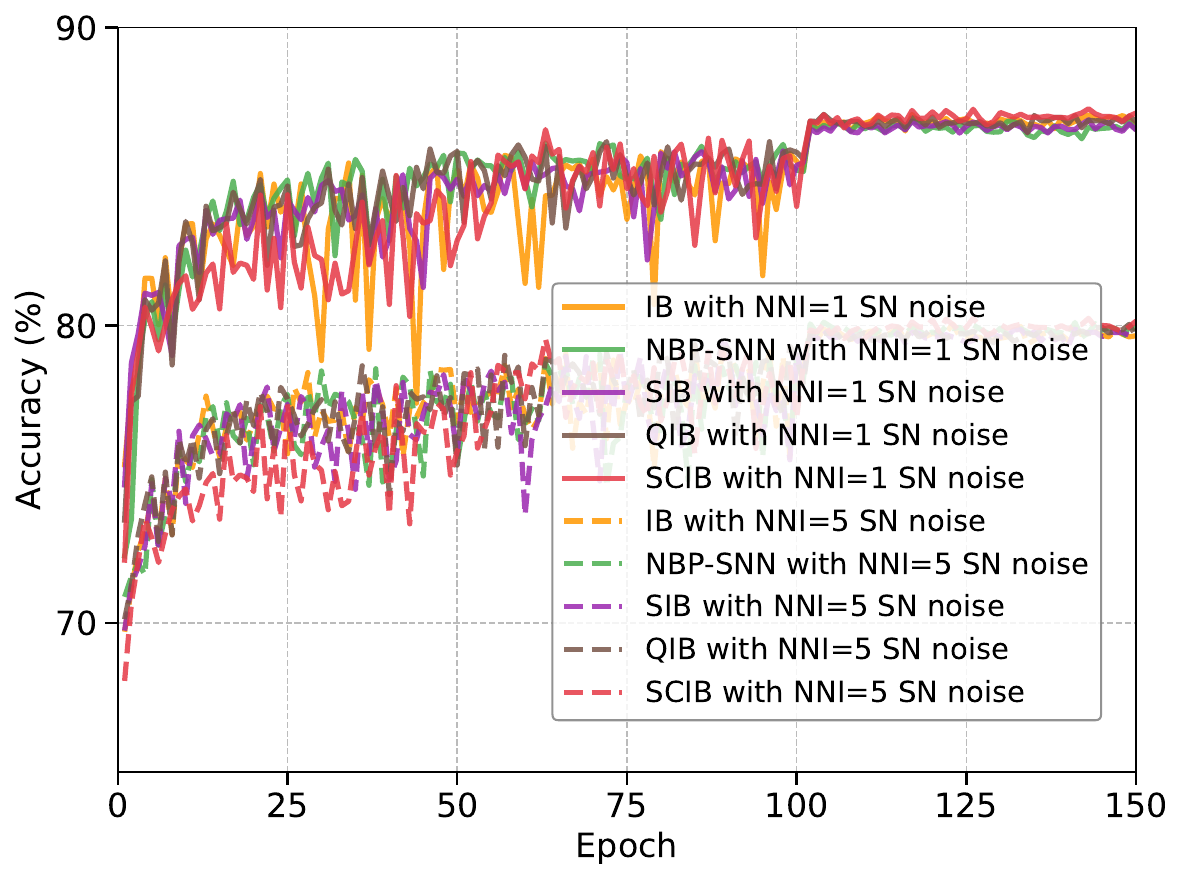}
        \caption{Fashion-MNIST}
        \label{fig:sn_fashion}
    \end{subfigure}
    \hfill
    \begin{subfigure}[b]{0.32\textwidth}
        \centering
        \includegraphics[width=\linewidth]{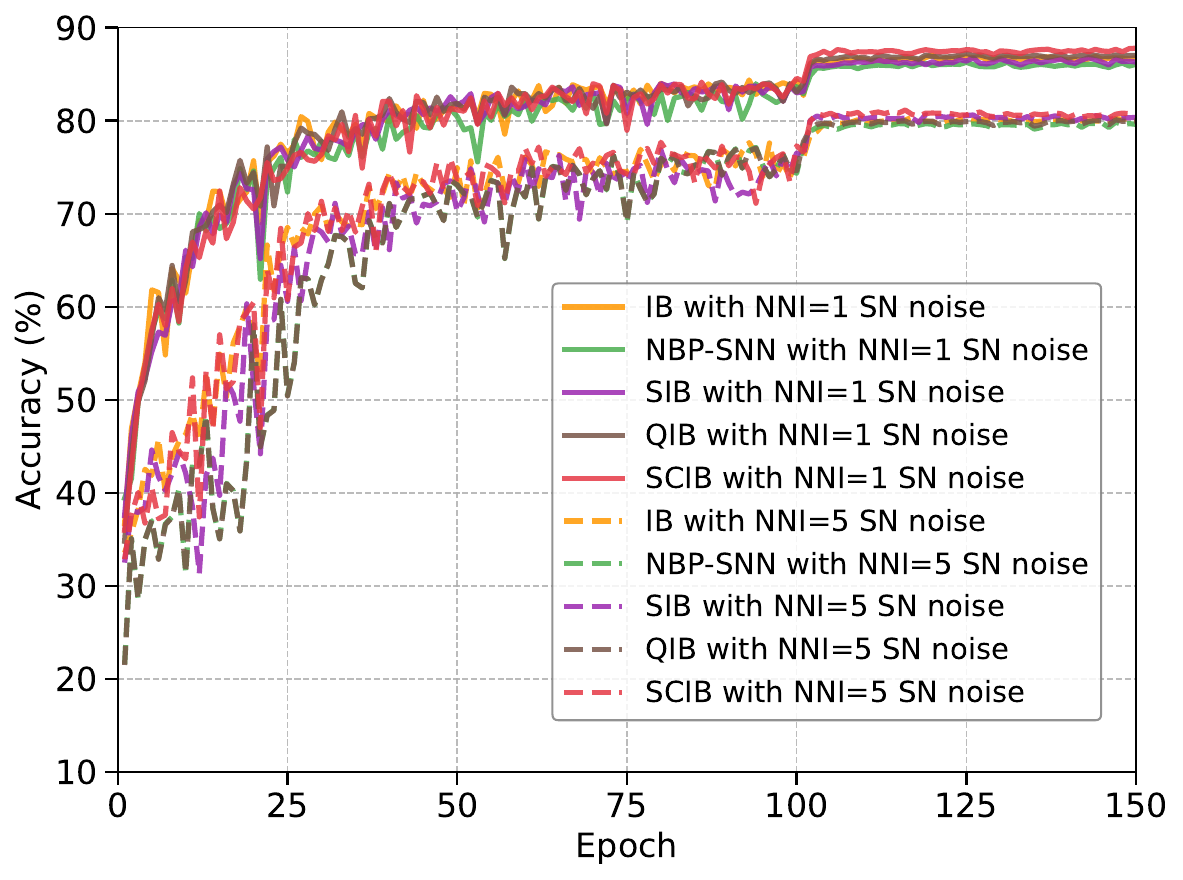}
        \caption{CIFAR-10}
        \label{fig:sn_cifar10}
    \end{subfigure}
    \hfill
    \begin{subfigure}[b]{0.32\textwidth}
        \centering
        \includegraphics[width=\linewidth]{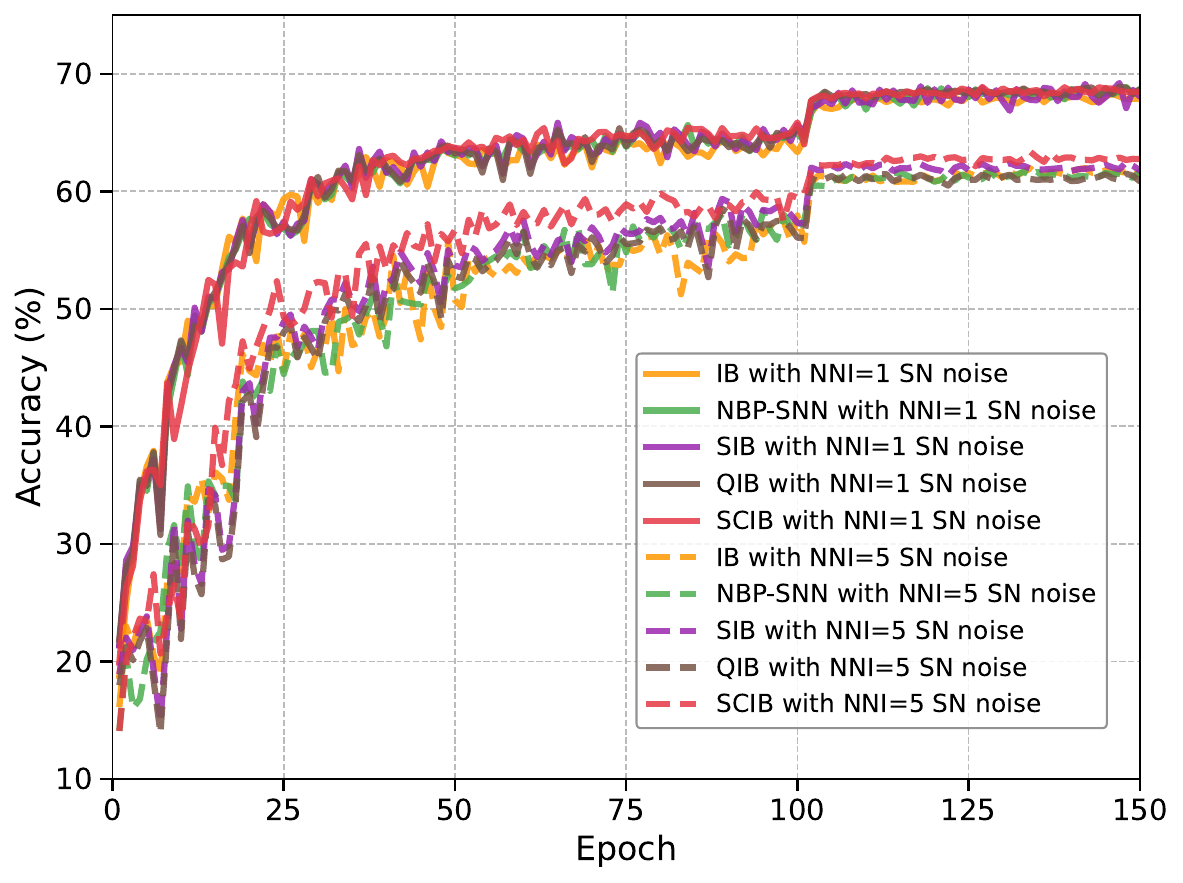}
        \caption{CIFAR-100}
        \label{fig:sn_cifar100}
    \end{subfigure}
    \caption{Training convergence curves of all methods under synaptic weight noise (SN). Solid and dashed lines correspond to NNI $= 1$ and NNI $= 5$, respectively.}
    \label{fig:sn}
\end{figure*}

When deploying SNNs on edge computing devices or mixed-signal neuromorphic chips, the non-ideal characteristics of physical devices introduce unavoidable low-level hardware noise. In this section, two representative internal perturbations are considered: synaptic weight noise and membrane potential noise, each tested at two intensity levels, NNI $= 1$ and NNI $= 5$. Figs.~\ref{fig:mpn} and~\ref{fig:sn} present the accuracy convergence curves of all methods on Fashion-MNIST, CIFAR-10, and CIFAR-100 under membrane potential noise (MPN) and synaptic noise (SN), respectively. Solid and dashed lines correspond to NNI $= 1$ and NNI $= 5$. All models are trained for 200 epochs; since the accuracy stabilizes after approximately epoch 110, only the first 150 epochs are shown.

Membrane potential noise acts directly on the state of LIF neurons and perturbs spike firing timing at each timestep, making it more destructive than synaptic weight noise. As shown in Fig.~\ref{fig:mpn}, under moderate perturbation (NNI $= 1$), the performance of all IB-based methods is comparable, with SCIB essentially on par with the best baseline. For example, SCIB achieves 88.18\% on CIFAR-10, close to IB's 88.20\%. Under extreme perturbation (NNI $= 5$), the performance gap becomes larger: on CIFAR-10, IB drops from 88.20\% to 64.08\%, while SCIB maintains 64.45\%. On CIFAR-100, SCIB achieves the best results under both perturbation levels, with its lead becoming more evident as the noise intensity increases. This suggests that class-conditional compression helps preserve discriminative spike representations under noisy neuronal dynamics.

Synaptic weight noise primarily affects the weight update and state readout processes of physical synapses. As shown in Fig.~\ref{fig:sn}, SCIB achieves best or near-best results in most configurations. On CIFAR-100, the performance gap among QIB, SIB, and SCIB is small under moderate perturbation, whereas the differences become more pronounced under extreme perturbation. When NNI increases to 5, SCIB achieves 62.98\% on CIFAR-100, approximately 0.9 percentage points above the second-best method. These results indicate that the benefit of conditional compression becomes more apparent when high-intensity synaptic noise accumulates in deep spike representations.

\subsection{Adversarial robustness}

To evaluate SCIB under gradient-based adversarial perturbations, this section compares SCIB with Vanilla SNN and FEEL-SNN~\cite{FEEL2024Xu} on CIFAR-10 and CIFAR-100. The attack methods include FGSM~\cite{Explaining2014Goodfellow}, PGD~\cite{Towards2018Madry}, and BIM~\cite{Adversarial2017Kurakin}, with perturbation budgets $\epsilon\in\{4/255,\; 8/255\}$ and timesteps $T=4$ and $T=8$. It should be particularly noted that SCIB is not explicitly optimized with adversarial examples during training; this section examines whether its conditional compression mechanism can provide implicit robustness gains in adversarial scenarios. The classification accuracy of all methods under different attack configurations is reported in Table~\ref{tab:adversarial}.

\begin{table*}[t]
\centering
\caption{Classification accuracy (\%) under adversarial attacks with perturbation budgets $\epsilon \in \{4/255, 8/255\}$. SCIB does not employ adversarial training. \textbf{Bold} and \underline{underline} denote the best and second-best results, respectively.}
\label{tab:adversarial}
\begin{tabular}{cccccccccc}
\hline
\multirow{2}{*}{Timestep} & 
\multirow{2}{*}{Dataset} & 
\multirow{2}{*}{Method} & 
\multirow{2}{*}{Clean} & 
\multicolumn{3}{c}{$\epsilon=4/255$} & 
\multicolumn{3}{c}{$\epsilon=8/255$} \\
\cline{5-10}
& & & & FGSM & PGD & BIM & FGSM & PGD & BIM \\
\hline
\multirow{6}{*}{$T=4$}
& \multirow{3}{*}{CIFAR-10}
    & Vanilla  & \textbf{92.94} & 19.63 & 0.78 & 0.67 & 8.29 & 0.02 & 0.01 \\
&   & FEEL-SNN & 88.07 & \underline{39.17} & \underline{21.33} & \underline{21.10} & \underline{16.19} & \underline{1.73} & \underline{1.38} \\
&   & SCIB     & \underline{89.69} & \textbf{52.08} & \textbf{47.58} & \textbf{47.73} & \textbf{37.02} & \textbf{26.65} & \textbf{26.04} \\
\cline{2-10}
& \multirow{3}{*}{CIFAR-100}
    & Vanilla  & \textbf{73.80} & 6.99 & 0.38 & 0.36 & 3.64 & 0.01 & 0.00 \\
&   & FEEL-SNN & 63.08 & \textbf{19.51} & \underline{9.08} & \underline{8.74} & \underline{8.27} & \underline{0.94} & \underline{0.76} \\
&   & SCIB     & \underline{64.73} & \underline{14.08} & \textbf{12.62} & \textbf{12.42} & \textbf{9.12} & \textbf{6.29} & \textbf{6.37} \\
\hline
\multirow{6}{*}{$T=8$}
& \multirow{3}{*}{CIFAR-10}
    & Vanilla  & \textbf{93.53} & 19.05 & 0.56 & 0.52 & 7.24 & 0.02 & 0.00 \\
&   & FEEL-SNN & 87.31 & \underline{39.70} & \underline{17.67} & \underline{17.71} & \underline{20.67} & \underline{1.60} & \underline{1.52} \\
&   & SCIB     & \underline{87.73} & \textbf{51.03} & \textbf{44.88} & \textbf{45.10} & \textbf{26.74} & \textbf{16.09} & \textbf{15.24} \\
\cline{2-10}
& \multirow{3}{*}{CIFAR-100}
    & Vanilla  & \textbf{74.33} & 7.43 & 0.47 & 0.47 & 3.88 & 0.02 & 0.00 \\
&   & FEEL-SNN & \underline{61.68} & \underline{15.42} & \underline{5.73} & \underline{5.57} & \underline{7.05} & \underline{0.51} & \underline{0.50} \\
&   & SCIB     & 61.53 & \textbf{25.39} & \textbf{21.94} & \textbf{21.65} & \textbf{11.91} & \textbf{7.55} & \textbf{7.18} \\
\hline
\end{tabular}
\end{table*}

Vanilla SNN without any defense mechanism collapses to near-zero accuracy under iterative PGD and BIM attacks, especially at $\epsilon=8/255$, confirming that standard SNN training provides limited inherent resistance to strong gradient-based adversarial perturbations. On CIFAR-10 with $T=4$ and $\epsilon=4/255$, SCIB improves over Vanilla SNN by 46.80 and 47.06 percentage points under PGD and BIM, respectively.

Compared with FEEL-SNN, SCIB shows stronger robustness under most iterative attacks. On CIFAR-10 with $T=4$, SCIB exceeds FEEL-SNN by 26.25 and 26.63 percentage points under PGD and BIM at $\epsilon=4/255$, and by 24.92 and 24.66 percentage points at $\epsilon=8/255$. A similar trend is observed at $T=8$, where SCIB consistently achieves the best adversarial accuracy on CIFAR-10 across all attack settings. On CIFAR-100, SCIB also outperforms FEEL-SNN under all PGD and BIM configurations, although FEEL-SNN is better under the single-step FGSM attack at $T=4$ and $\epsilon=4/255$.

These results suggest that SCIB can improve robustness under the evaluated gradient-based attacks, particularly iterative attacks. Since SCIB is not adversarially trained, these results should be interpreted as implicit robustness gains from representation-level regularization rather than evidence that SCIB dominates dedicated adversarial-training defenses.

\subsection{Energy consumption}

In the deployment of edge computing devices and mixed-signal neuromorphic chips, computational complexity and dynamic power consumption are important metrics. SNNs rely on an event-driven mechanism, transmitting binary spikes to postsynaptic neurons only when the membrane potential exceeds the firing threshold. Therefore, physical energy consumption is directly related to the frequency of spike events.

To provide an algorithm-level proxy for inference efficiency, we construct a analytical computation-energy model based on Synaptic Operations (SynOps). For a hidden layer $l$, let the output feature map size be $H_l \times W_l \times C_l^{\mathrm{out}}$, the convolutional kernel size be $k_l$, and the number of input channels be $C_l^{\mathrm{in}}$. The fan-in of each neuron is $N_l = k_l^2 C_l^{\mathrm{in}}$. Defining the average total number of spike firings of a single neuron in layer $l$ over the inference window as $f_l$, the total synaptic accumulation operations are
\begin{equation}
    \mathrm{AC}_{l} = H_{l} \cdot W_{l} \cdot 
    C_{l}^{\mathrm{out}} \cdot N_{l} \cdot f_{l}.
    \label{eq:ac}
\end{equation}

Following the commonly used 45-nm CMOS operation-energy estimates~\cite{Computing2014Horowitz}, a 32-bit floating-point MAC operation costs approximately $E_{\mathrm{MAC}} = 4.6\,\mathrm{pJ}$, while an AC operation costs $E_{\mathrm{AC}} = 0.9\,\mathrm{pJ}$. This analytical model accounts only for computation energy associated with MAC/AC operations.
\begin{equation}
    \mathrm{MAC}_{1} = H_{1} \cdot W_{1} \cdot 
    C_{1}^{\mathrm{out}} \cdot N_{1} \cdot T.
    \label{eq:mac}
\end{equation}

All subsequent layers are driven by binary spikes and execute AC operations. The operation-level dynamic inference-energy proxy is estimated as
\begin{equation}
    E_{\mathrm{total}} = E_{\mathrm{MAC}} \cdot 
    \mathrm{MAC}_{1} + E_{\mathrm{AC}} \cdot 
    \sum_{l=2}^{L} \mathrm{AC}_{l}
    \label{eq:energy}
\end{equation}

From this model, SNN inference energy is linearly related to layer-wise firing rates. SCIB reduces spike activity by regularizing label-conditioned input-specific dependence, thereby lowering SynOps in deeper layers under this analytical model. Taking CIFAR-10 as an example, the per-layer spike firing statistics and synaptic operations of SCIB are reported in Table~\ref{tab:synops}.

\begin{table}
\centering
\caption{Per-layer spike firing statistics and synaptic operations of SCIB on CIFAR-10. M denotes MAC operations; A denotes AC operations.}
\label{tab:synops}
\begin{tabular}{ccccc}
\hline
Layer  & Neurons & Spikes    & Rate   & SynOps          \\ 
\hline
Conv\#1 & 65536  & 131859.7  & 2.0120 & $1.8\times10^{6}$ (M) \\
Conv\#2 & 65536  & 142810.9  & 2.1791 & $7.6\times10^{7}$ (A) \\
Conv\#3 & 32768  & 74278.7   & 2.2668 & $4.1\times10^{7}$ (A) \\
Conv\#4 & 32768  & 69054.9   & 2.1074 & $8.6\times10^{7}$ (A) \\
Conv\#5 & 16384  & 37739.2   & 2.3034 & $4.0\times10^{7}$ (A) \\
Conv\#6 & 16384  & 40166.4   & 2.4516 & $8.7\times10^{7}$ (A) \\
Conv\#7 & 16384  & 43500.8   & 2.6551 & $9.3\times10^{7}$ (A) \\
FC1    & 1024   & 2367.6    & 2.3121 & $1.1\times10^{7}$ (A) \\
\hline
\end{tabular}
\end{table}

As shown in Table~\ref{tab:synops}, except for Conv\#1, all subsequent layers perform low-power AC operations. The average firing rates of deeper layers such as Conv\#6 and Conv\#7 are controlled at approximately 2.45 and 2.65, respectively.

\begin{figure}
    \centering
    \includegraphics[width=\columnwidth]{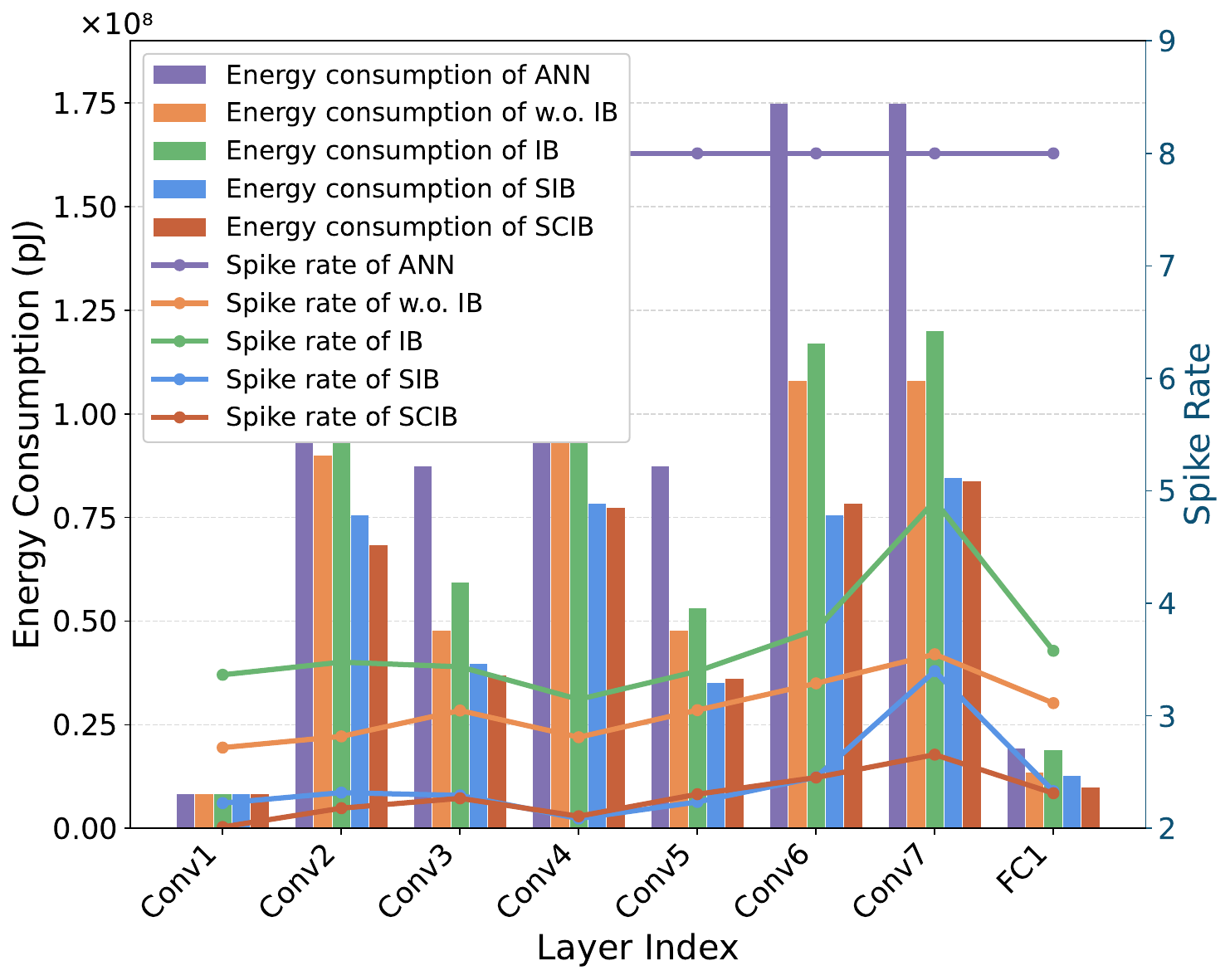}
    \caption{Comparison of average spike firing rate and per-layer energy consumption during single-sample inference.}
    \label{fig:energy}
\end{figure}

Fig.~\ref{fig:energy} presents the average spike firing rate and per-layer energy consumption. Across all methods, Conv\#1 has the same energy consumption because it directly processes analog inputs through dense MAC operations. From Conv\#2 onward, the differences among methods become evident. Compared with the baselines, SCIB maintains lower firing rates in most layers, especially in deeper convolutional layers.

By accumulating layer-wise operation-energy estimates, the total SynOps-based single-sample inference-energy proxy is approximately $9.02\times10^{8}\,\mathrm{pJ}$ for ANN, $5.31\times10^{8}\,\mathrm{pJ}$ for w.o.~IB, $6.11\times10^{8}\,\mathrm{pJ}$ for IB, $4.10\times10^{8}\,\mathrm{pJ}$ for SIB, and $3.99\times10^{8}\,\mathrm{pJ}$ for SCIB. Overall, under this analytical SynOps-based model, SCIB reduces the estimated operation-level inference energy by approximately $55.8\%$ compared with ANN and $24.9\%$ compared with w.o.~IB.

\subsection{Further analysis}
\subsubsection{Training evolution of the learnable class-conditional prior}

\begin{figure*}[t]
    \centering
    \begin{subfigure}[b]{0.24\textwidth}
        \centering
        \includegraphics[width=\linewidth]{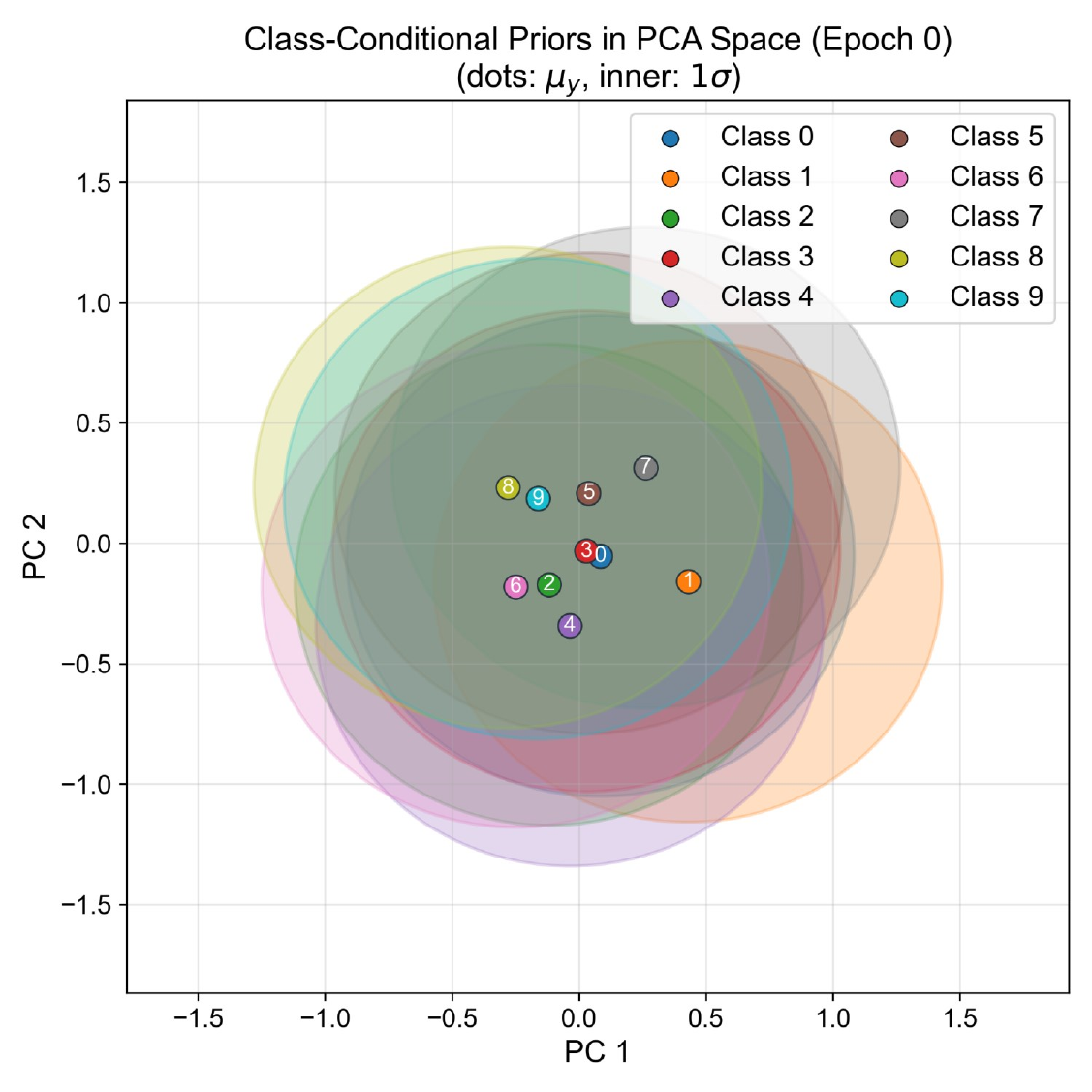}
        \caption*{Epoch 0}
    \end{subfigure}
    \hfill
    \begin{subfigure}[b]{0.24\textwidth}
        \centering
        \includegraphics[width=\linewidth]{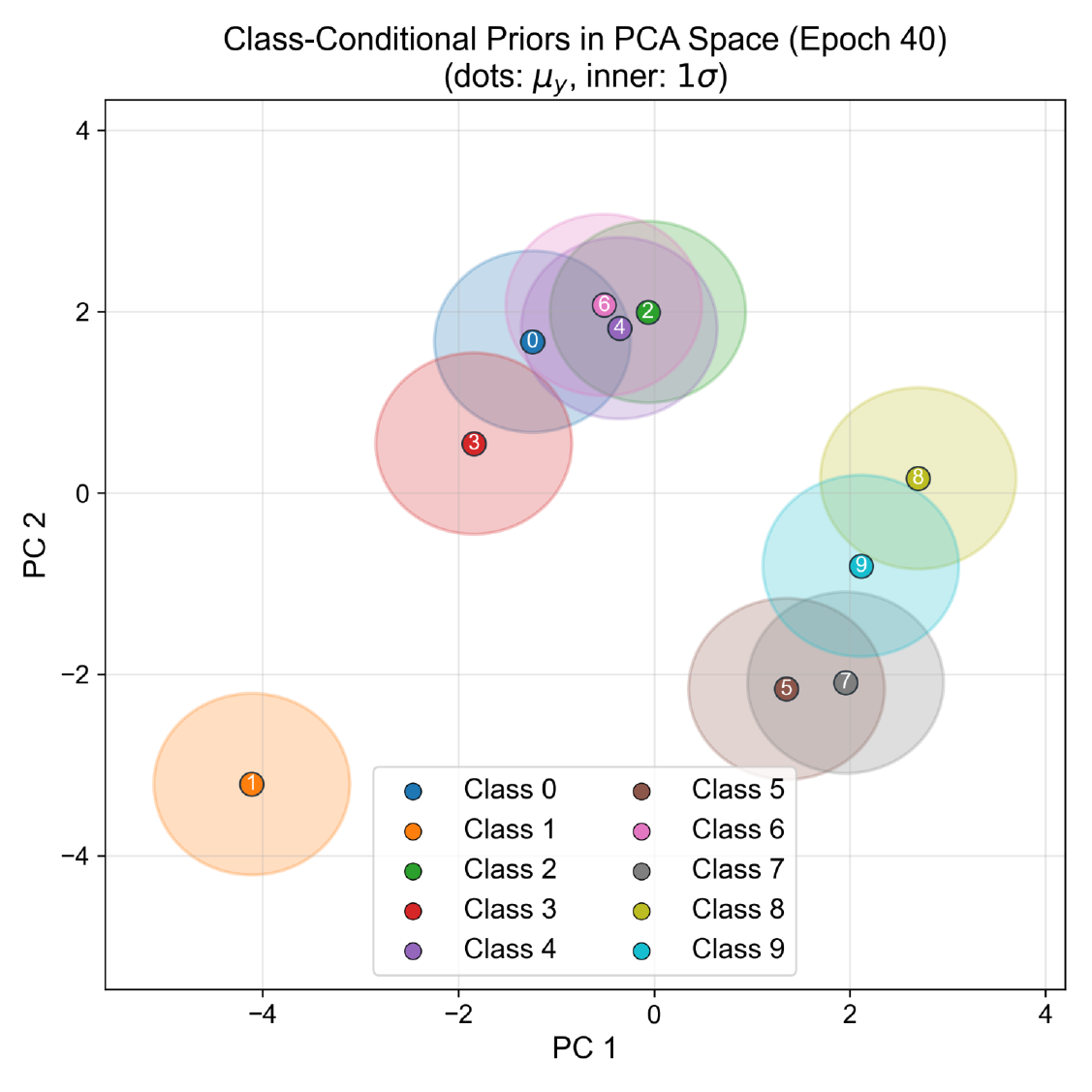}
        \caption*{Epoch 40}
    \end{subfigure}
    \hfill
    \begin{subfigure}[b]{0.24\textwidth}
        \centering
        \includegraphics[width=\linewidth]{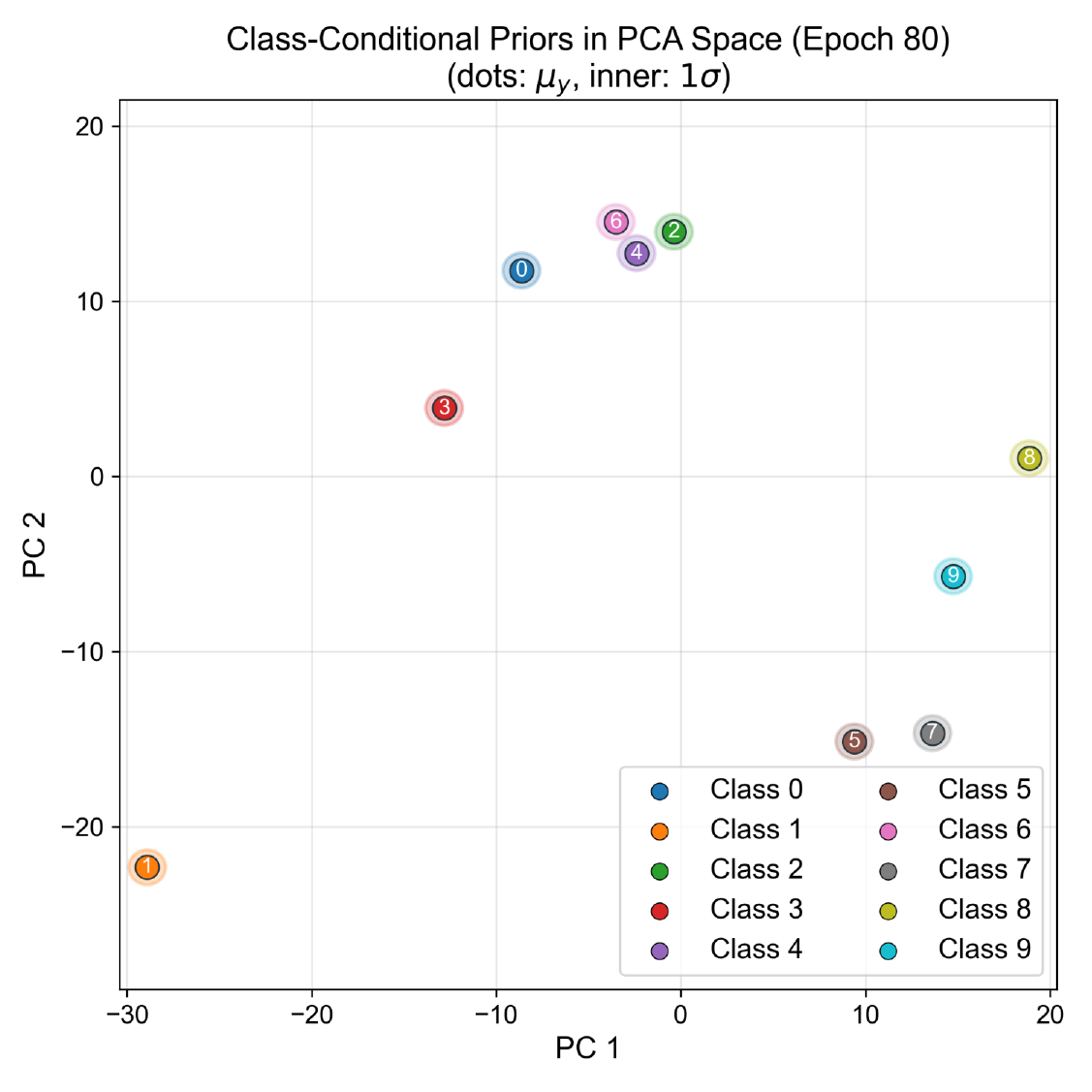}
        \caption*{Epoch 80}
    \end{subfigure}
    \hfill
    \begin{subfigure}[b]{0.24\textwidth}
        \centering
        \includegraphics[width=\linewidth]{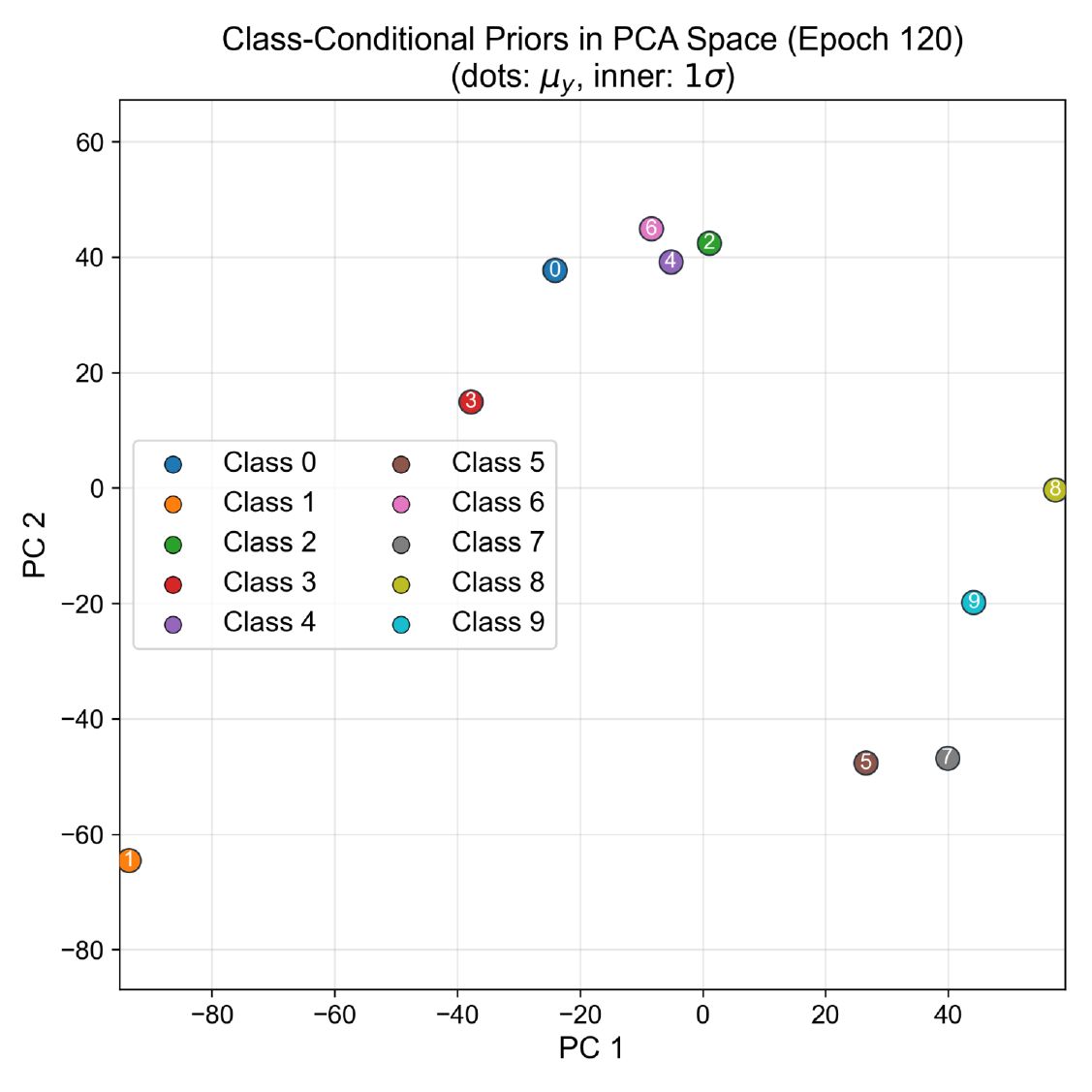}
        \caption*{Epoch 120}
    \end{subfigure}

    \vspace{0.5em}
    \begin{subfigure}[b]{\textwidth}
        \centering
        \caption*{(a) PCA projection evolution of class-conditional prior means $\mu_k$}
    \end{subfigure}

    \vspace{0.8em}

    \begin{subfigure}[b]{0.24\textwidth}
        \centering
        \includegraphics[width=\linewidth]{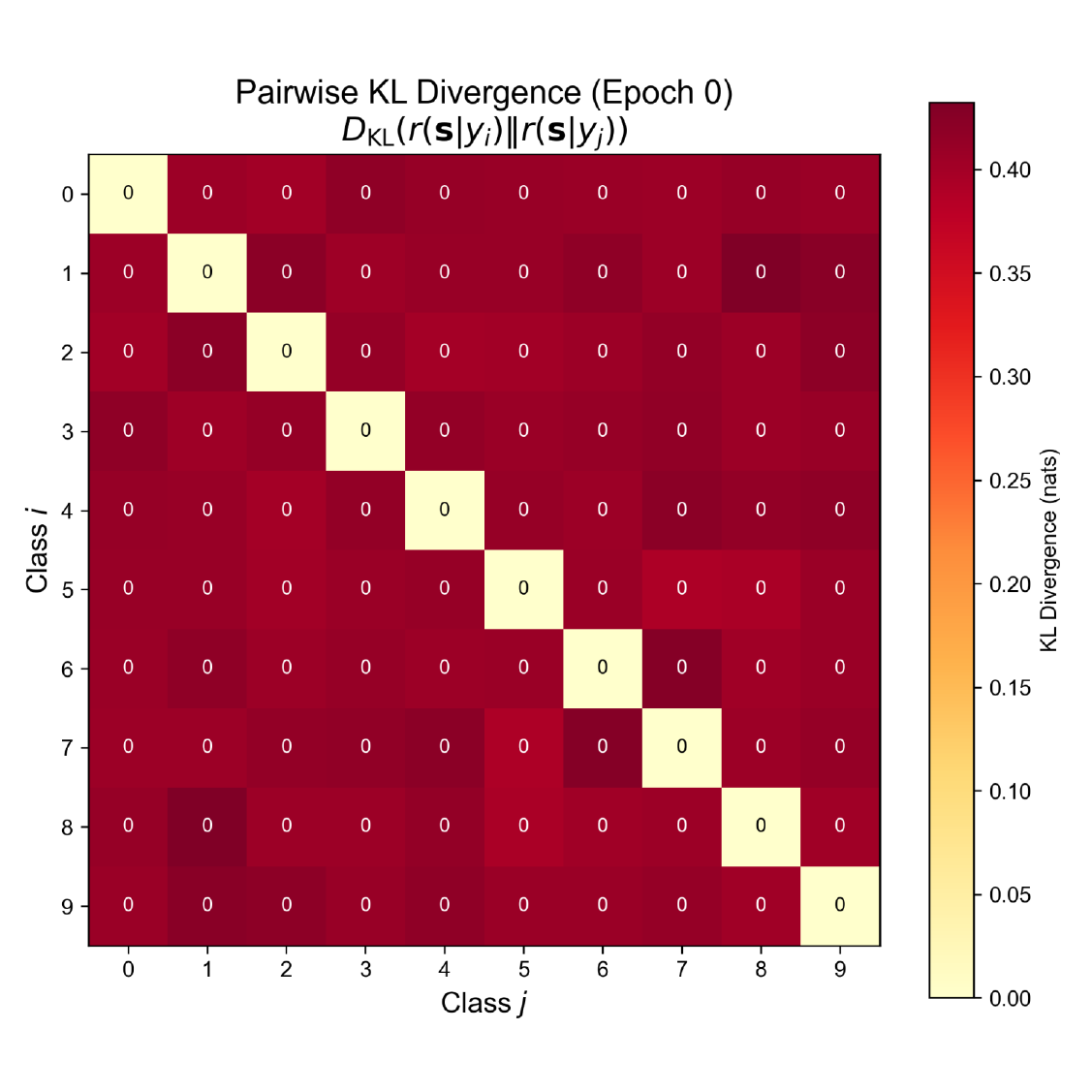}
        \caption*{Epoch 0}
    \end{subfigure}
    \hfill
    \begin{subfigure}[b]{0.24\textwidth}
        \centering
        \includegraphics[width=\linewidth]{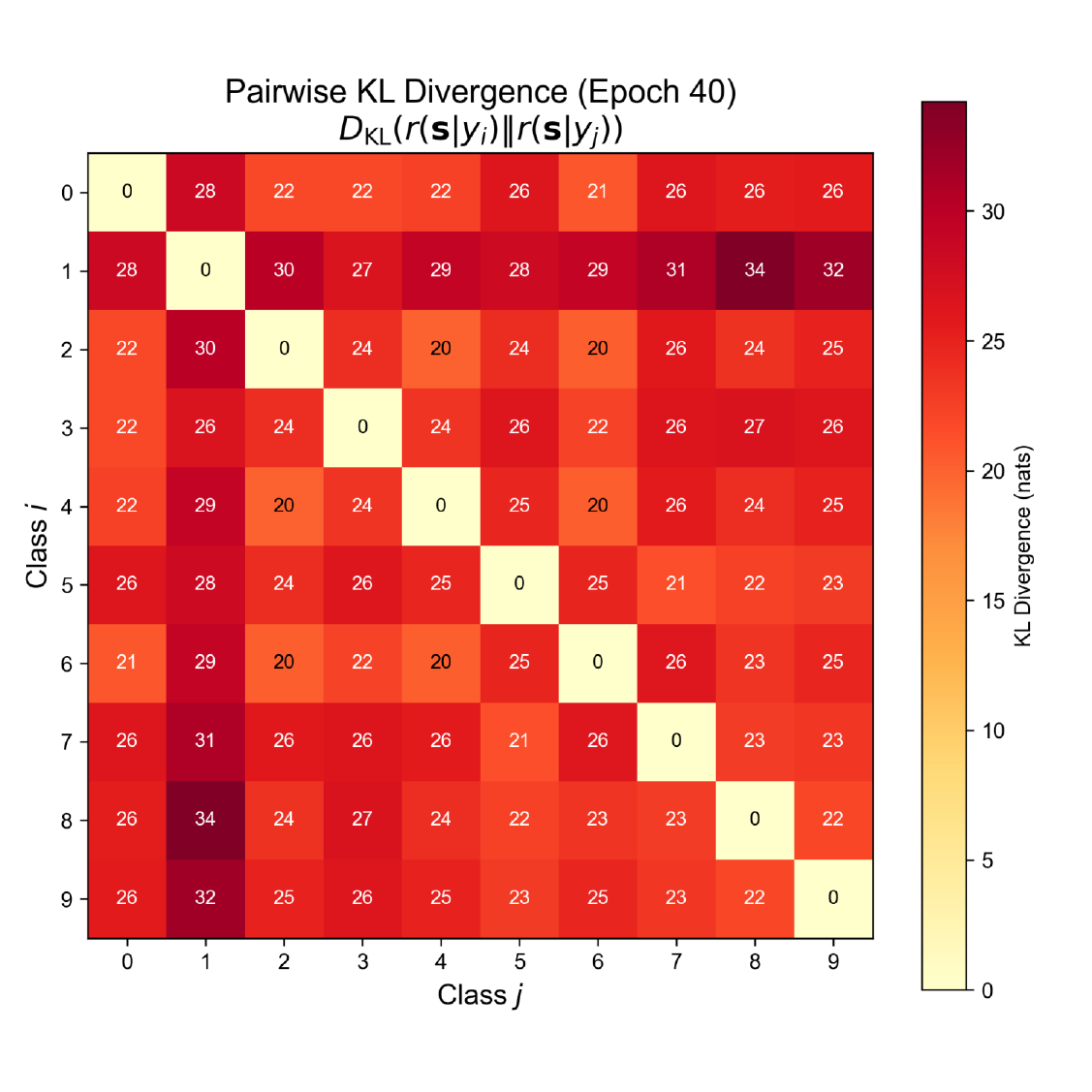}
        \caption*{Epoch 40}
    \end{subfigure}
    \hfill
    \begin{subfigure}[b]{0.24\textwidth}
        \centering
        \includegraphics[width=\linewidth]{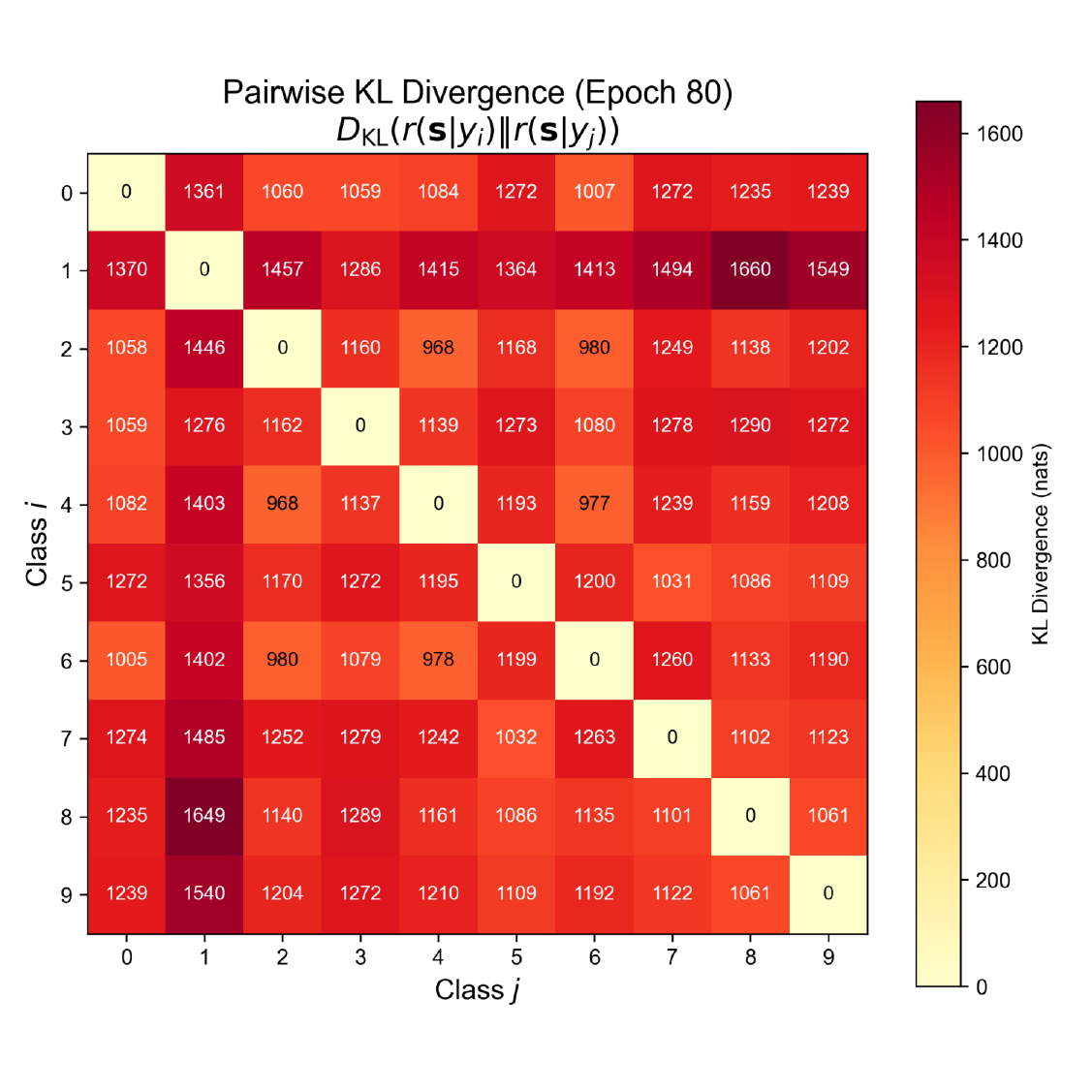}
        \caption*{Epoch 80}
    \end{subfigure}
    \hfill
    \begin{subfigure}[b]{0.24\textwidth}
        \centering
        \includegraphics[width=\linewidth]{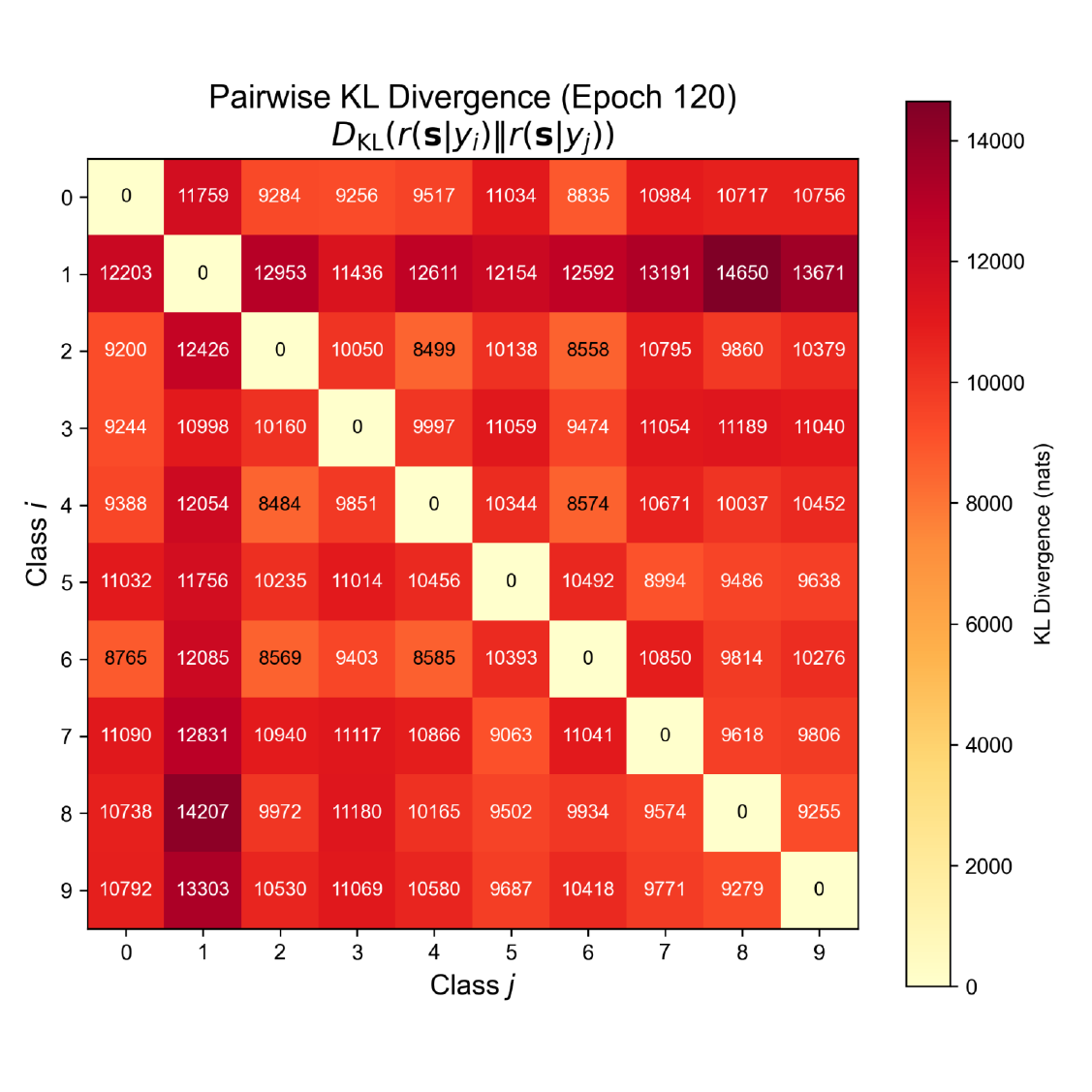}
        \caption*{Epoch 120}
    \end{subfigure}

    \vspace{0.5em}
    \begin{subfigure}[b]{\textwidth}
        \centering
        \caption*{(b) Inter-class KL divergence heatmap evolution}
    \end{subfigure}

    \caption{Training evolution of the class-conditional prior distribution $r(s|y)$. Row~(a): PCA projections of prior mean vectors; colored dots mark mean positions and $1\sigma$ confidence ellipses indicate the local coverage of each class prior. Row~(b): KL divergence heatmaps between class-conditional priors.}
    \label{fig:prior_evo}
\end{figure*}

This section further conducts a visualization analysis of the parameter evolution of the learnable class-conditional prior distribution proposed in Section~\ref{sec:class_conditional_prior}, examining how the class-conditional regularizer shapes the learned prior structure in the representation space. Specifically, a VGG-9 model is trained on Fashion-MNIST, and the mean vectors and variance vectors of the 10 class prior distributions are extracted at four training stages---Epoch 0, 40, 80, and 120---and visualized from two complementary perspectives: the global topological structure of the latent space and the inter-class statistical distances. Results are shown in Fig.~\ref{fig:prior_evo}.

Fig.~\ref{fig:prior_evo}(a) shows the spatial distribution of each class prior mean vector projected onto a two-dimensional plane via Principal Component Analysis (PCA), where colored dots mark the mean positions and $1\sigma$ confidence ellipses represent the local coverage of each class prior. At the initial training stage (Epoch 0), the prior means of all classes are densely clustered within a small region near the origin ($\pm 0.5$), with confidence ellipses heavily overlapping and classes nearly indistinguishable in the latent space. As training progresses, the class prior centers spread out in the latent space driven by the class-conditional KL regularization and the classification objective, while the confidence ellipses exhibit a pronounced shrinkage trend: by Epoch 80, the confidence ellipses appear largely separated in the PCA projection; by Epoch 120, the inter-center distances further increase and the ellipses become more compact, indicating that the prior parameters have largely converged.

Furthermore, the spatial topology learned by the prior distributions exhibits a clustering pattern highly consistent with class semantics. The top-wear categories in Fashion-MNIST---Class~0 (T-shirt), Class~2 (Pullover), Class~4 (Coat), and Class~6 (Shirt)---consistently maintain a nearest-neighbor clustering relationship in the prior space; the footwear categories Class~5 (Sandal) and Class~7 (Sneaker) form a neighboring pair; Class~1 (Trouser), whose morphological features differ from the top-wear categories, remains relatively separated from those categories throughout training. This suggests that the learnable prior mechanism, together with the classification objective, encourages a class-structured latent topology that is partly consistent with semantic similarity between categories.

Fig.~\ref{fig:prior_evo}(b) provides a complementary quantitative view of the evolution process from the perspective of statistical distances. At the initial training stage (Epoch 0), all off-diagonal elements of the KL divergence matrix are close to zero ($< 0.5$ nats), indicating that the class priors are statistically nearly indistinguishable. As training iterates, the off-diagonal elements exhibit a generally increasing trend: the KL divergence between Class~1 (Trouser) and most other classes remains among the largest values across training stages, corroborating the spatial observations in the PCA visualization; the KL divergences between intra-cluster class pairs within the top-wear cluster are relatively low, quantitatively supporting the topological property that semantically similar classes tend to maintain nearest-neighbor relationships in the prior space.

Integrating both levels of analysis, the learnable class-conditional prior mechanism exhibits a complete evolutionary trajectory during training, from random initialization to class-specific spatial partitioning: confidence ellipses progressively shrink from heavily overlapping to more compact regions; prior centers move from origin-clustered to more spatially separated configurations with increasing pairwise KL divergences; and visually similar classes tend to form neighboring groups in the prior space. This visualization provides empirical evidence that the conditional regularization term is associated with more class-structured latent priors, although the observed separation should not be interpreted as a theoretical guarantee.

\subsubsection{Sensitivity analysis of compression strength hyperparameter \texorpdfstring{$\beta$}{beta}}

In the optimization objective of the SCIB framework, the hyperparameter $\beta$ determines the compression strength of the information bottleneck, controlling the trade-off between maintaining predictive performance and regularizing label-conditioned input-specific information. When $\beta$ approaches zero, the compression constraint relaxes and the model tends to retain more original input information to prioritize prediction accuracy; as $\beta$ increases, the compression constraint strengthens and the model is forced to discard more input-specific information---potentially achieving superior robustness in noisy environments, but also risking the loss of useful discriminative information due to over-compression. This section conducts a systematic sensitivity analysis of $\beta$ on CIFAR-10 using the VGG-9 architecture.

\begin{figure*}
    \centering
    \begin{subfigure}[b]{0.48\textwidth}
        \centering
        \includegraphics[width=\linewidth]{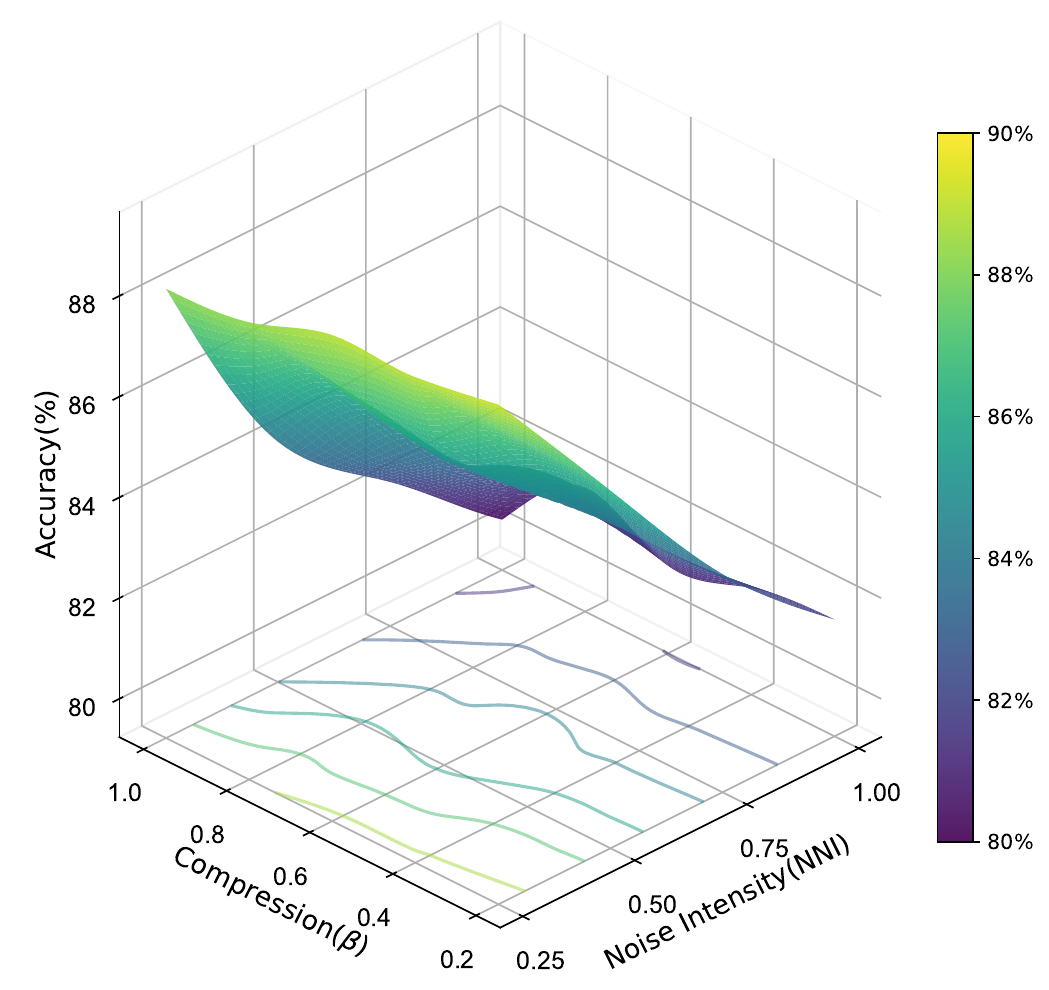}
        \caption{Gaussian noise}
        \label{fig:beta_gaussian}
    \end{subfigure}
    \hfill
    \begin{subfigure}[b]{0.48\textwidth}
        \centering
        \includegraphics[width=\linewidth]{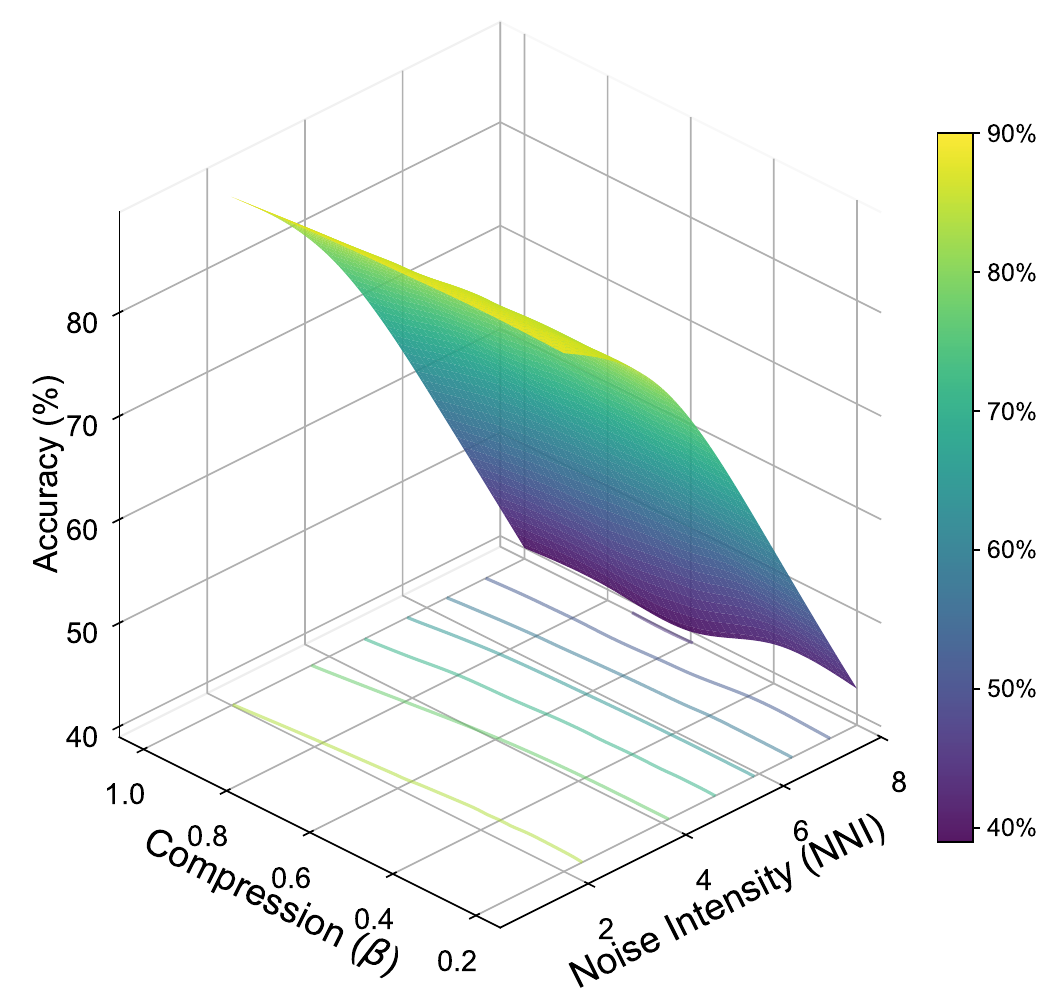}
        \caption{membrane potential noise}
        \label{fig:beta_mpn}
    \end{subfigure}
    \caption{Sensitivity analysis of the compression strength hyperparameter $\beta$.}
    \label{fig:beta_sensitivity}
\end{figure*}

For the Gaussian noise experiments, NNI is set to 0.25, 0.50, 0.75, and 1.00; for the membrane potential noise experiments, NNI is set to 2, 4, 6, and 8. At each noise intensity level, $\beta$ takes values of 0.2, 0.4, 0.6, 0.8, and 1.0, forming $4 \times 5 = 20$ cross-experimental configurations. The experimental results are shown in Fig.~\ref{fig:beta_sensitivity}.

Fig.~\ref{fig:beta_sensitivity}(a) presents the $\beta$--NNI--accuracy response surface under Gaussian noise. The surface is relatively flat, indicating that model performance is not highly sensitive to $\beta$ under input-level Gaussian noise: under low-noise conditions, the performance differences across different $\beta$ values are minimal; as noise intensity increases, smaller $\beta$ values generally maintain higher accuracy, but moderately increasing $\beta$ at higher noise intensities can instead achieve a local optimum, suggesting that strengthening the compression constraint may help reduce perturbation-sensitive information when noise interference is stronger. Fig.~\ref{fig:beta_sensitivity}(b) presents the response surface under membrane potential noise, where increasing $\beta$ leads to a monotonically decreasing accuracy trend at all noise intensity levels, with this trend becoming more pronounced under high-noise conditions.

Synthesizing the observations across both noise paradigms, the optimal $\beta$ is not fixed but requires adaptive adjustment according to the noise type and intensity: under Gaussian noise, $\beta$ and performance exhibit a non-monotonic relationship, where moderate compression is beneficial under strong noise; under membrane potential noise, $\beta$ and performance exhibit a monotonically decreasing relationship, and smaller compression strengths should be preferred. This analysis reveals the scenario-dependent nature of $\beta$ selection in the SCIB framework, providing systematic experimental evidence for hyperparameter tuning under different noise environments.

\section{Conclusion}

Spiking Neural Networks (SNNs) provide an energy-efficient computing paradigm for neuromorphic perception because of event-driven processing and sparse spike communication, but robustness remains a critical challenge. To address this issue, this paper proposes SCIB, a Spiking Conditional Information Bottleneck framework for robust and energy-efficient SNN representation learning. Motivated by the empirical observation that standard-trained SNNs may retain substantial or even redundant input-related information after their label-predictive information becomes stable, SCIB formulates robustness enhancement as a representation-level conditional compression problem. Unlike conventional IB-based methods that directly compress the mutual information $I(X;S)$, SCIB adopts a variational upper bound of the conditional mutual information $I(X;S|Y)$ as the compression regularizer, thereby encouraging class-conditioned compactness while relying on the supervised objective to preserve label-relevant predictive information. To make this objective tractable and suitable for spike-based learning, we derive a variational upper bound for conditional compression and introduce a learnable class-conditional prior, which provides class-specific distributional anchors and improves the expressiveness of the variational prior compared with a label-agnostic standard Gaussian. 

Experiments demonstrate that SCIB improves robustness under external environmental noise, neuromorphic hardware noise, and adversarial attacks. Across most noise configurations, SCIB outperforms standard SNNs and existing IB-based baselines, with advantages that become more evident under stronger perturbations. In addition, the SynOps-based energy analysis shows that SCIB reduces redundant spike firing and lowers single-sample inference energy consumption by approximately 24.9\% compared with the unconstrained SNN baseline, indicating that conditional compression regularization can jointly improve robustness and reduce operation-level SynOps energy estimates in the evaluated settings.

Several limitations remain for future investigation. First, SCIB does not explicitly optimize against adversarial examples, and thus its robustness under certain single-step attacks may still be limited. Second, the current class-conditional prior assigns an independent Gaussian component to each class, which may require more careful design for datasets with many fine-grained categories. Future work will explore more sample-efficient and hierarchical prior structures, investigate the complementary combination of SCIB with adversarial training, and validate the proposed framework on physical neuromorphic hardware.

\printcredits

\bibliographystyle{unsrtnat}

\bibliography{main.bib}

@ARTICLE{System2020Zhang,
  author  = {Zhang, Y. and Qu, P. and Ji, Y. and Zhang, W. and Gao, G. and Wang, G. and Shi, L.},
  title   = {A system hierarchy for brain-inspired computing},
  journal = {Nature}, 
  volume  = {586},
  year    = {2020},
  pages   = {378-384}
}

@ARTICLE{Brain2022Mehonic,
  author  = {Mehonic, A. and Kenyon, A. J.},
  title   = {Brain-inspired computing needs a master plan},
  journal = {Nature}, 
  volume  = {604},
  year    = {2022},
  pages   = {255-260}
}

@misc{Explaining2014Goodfellow,
  author  = {Goodfellow, I. J. and Shlens, J. and Szegedy, C.},
  title   = {Explaining and harnessing adversarial examples},
  howpublished = {arXiv preprint arXiv:1412.6572},
  year    = {2014}
}

@ARTICLE{Networks1997Maass,
  author  = {Maass, W.},
  title   = {Networks of spiking neurons: the third generation of neural network models},
  journal = {Neural Netw.}, 
  volume  = {10},
  year    = {1997},
  pages   = {1659-1671}
}

@ARTICLE{Towards2019Roy,
  author  = {Roy, K. and Jaiswal, A. and Panda, P.},
  title   = {Towards spike-based machine intelligence with neuromorphic computing},
  journal = {Nature}, 
  volume  = {575},
  year    = {2019},
  pages   = {607-617}
}

@ARTICLE{Loihi2018Davies,
  author  = {Davies, M. and Srinivasa, N. and Lin, T. H. and Chinya, G. and Cao, Y. and Choday, S. H. and Wang, H.},
  title   = {Loihi: A neuromorphic manycore processor with on-chip learning},
  journal = {IEEE Micro}, 
  volume  = {38},
  year    = {2018},
  pages   = {82-99}
}

@ARTICLE{Truenorth2015Akopyan,
  author  = {Akopyan, F. and Sawada, J. and Cassidy, A. and Alvarez-Icaza, R. and Arthur, J. and Merolla, P. and Modha, D. S.},
  title   = {Truenorth: Design and tool flow of a 65 mw 1 million neuron programmable neurosynaptic chip},
  journal = {IEEE Trans. Comput.-Aided Des. Integr. Circuits Syst.}, 
  volume  = {34},
  year    = {2015},
  pages   = {1537-1557}
}

@ARTICLE{Tianjic2020Deng,
  author  = {Deng, L. and Wang, G. and Li, G. and Li, S. and Liang, L. and Zhu, M. and Shi, L.},
  title   = {Tianjic: A unified and scalable chip bridging spike-based and continuous neural computation},
  journal = {IEEE J. Solid-State Circuits}, 
  volume  = {55},
  year    = {2020},
  pages   = {2228-2246}
}

@ARTICLE{Advancing2021Davies,
  author  = {Davies, M. and Wild, A. and Orchard, G. and Sandamirskaya, Y. and Guerra, G. A. F. and Joshi, P. and Risbud, S. R.},
  title   = {Advancing neuromorphic computing with loihi: A survey of results and outlook},
  journal = {Proc. IEEE}, 
  volume  = {109},
  year    = {2021},
  pages   = {911-934}
}

@ARTICLE{Opportunities2022Schuman,
  author  = {Schuman, C. D. and Kulkarni, S. R. and Parsa, M. and Mitchell, J. P. and Date, P. and Kay, B.},
  title   = {Opportunities for neuromorphic computing algorithms and applications},
  journal = {Nat. Comput. Sci.}, 
  volume  = {2},
  year    = {2022},
  pages   = {10-19}
}

@ARTICLE{Spatio2018Wu,
  author  = {Wu, Y. and Deng, L. and Li, G. and Zhu, J. and Shi, L.},
  title   = {Spatio-temporal backpropagation for training high-performance spiking neural networks},
  journal = {Front. Neurosci.}, 
  volume  = {12},
  year    = {2018},
  pages   = {331}
}

@CONFERENCE{Spiking2020Kim,
  author  = {Kim, S. and Park, S. and Na, B. and Yoon, S.},
  title   = {Spiking-yolo: spiking neural network for energy-efficient object detection},
  booktitle = {In Proc. AAAI Conf. Artif. Intell.}, 
  year    = {2020},
  pages   = {11270-11277}
}

@ARTICLE{Spiking2015Cao,
  author  = {Cao, Y. and Chen, Y. and Khosla, D.},
  title   = {Spiking deep convolutional neural networks for energy-efficient object recognition},
  journal = {Int. J. Comput. Vis.}, 
  volume  = {113},
  year    = {2015},
  pages   = {54-66}
}

@ARTICLE{Supervised2017Mostafa,
  author  = {Mostafa, H.},
  title   = {Supervised learning based on temporal coding in spiking neural networks},
  journal = {IEEE Trans. Neural Netw. Learn. Syst.}, 
  volume  = {29},
  year    = {2017},
  pages   = {3227-3235}
}

@CONFERENCE{Direct2019Wu,
  author  = {Wu, Y. and Deng, L. and Li, G. and Zhu, J. and Xie, Y. and Shi, L.},
  title   = {Direct training for spiking neural networks: Faster, larger, better},
  booktitle = {In Proc. AAAI Conf. Artif. Intell.}, 
  year    = {2019},
  pages   = {1311-1318}
}

@ARTICLE{Surrogate2019Neftci,
  author  = {Neftci, E. O. and Mostafa, H. and Zenke, F.},
  title   = {Surrogate gradient learning in spiking neural networks},
  journal = {IEEE Signal Process. Mag.}, 
  volume  = {36},
  year    = {2019},
  pages   = {51-63}
}

@misc{Comprehensive2019Sharmin,
  author  = {Sharmin, S. and Panda, P. and Sarwar, S. S. and Lee, C. and Ponghiran, W. and Roy, K.},
  title   = {A comprehensive analysis on adversarial robustness of spiking neural networks},
  howpublished = {arXiv preprint arXiv:1905.02704},
  year    = {2019}
}

@ARTICLE{BiCoSS2021Yang,
  author  = {Yang, S. and Wang, J. and Hao, X. and Li, H. and Wei, X. and Deng, B. and Loparo, K. A.},
  title   = {BiCoSS: Toward large-scale cognition brain with multigranular neuromorphic architecture},
  journal = {IEEE Trans. Neural Netw. Learn. Syst.}, 
  volume  = {33},
  year    = {2021},
  pages   = {2801-2815}
}

@ARTICLE{Exploring2021Liang,
  author  = {Liang, L. and Hu, X. and Deng, L. and Wu, Y. and Li, G. and Ding, Y. and Xie, Y.},
  title   = {Exploring adversarial attack in spiking neural networks with spike-compatible gradient},
  journal = {IEEE Trans. Neural Netw. Learn. Syst.}, 
  volume  = {34},
  year    = {2021},
  pages   = {2569-2583}
}

@CONFERENCE{Rate2023Bu,
  author  = {Bu, T. and Ding, J. and Hao, Z. and Yu, Z.},
  title   = {Rate gradient approximation attack threats deep spiking neural networks},
  booktitle = {In Proc. IEEE/CVF Conf. Comput. Vis. Pattern Recognit.}, 
  year    = {2023},
  pages   = {7896-7906}
}

@misc{Information2000Tishby,
  author  = {Tishby, N. and Pereira, F. C. and Bialek, W.},
  title   = {The information bottleneck method},
  howpublished = {arXiv preprint physics/0004057},
  year    = {2000}
}

@CONFERENCE{Deep2015Tishby,
  author  = {Tishby, N. and Zaslavsky, N.},
  title   = {Deep learning and the information bottleneck principle},
  booktitle = {In Proc. IEEE Inf. Theory Workshop (ITW)}, 
  year    = {2015},
  pages   = {1-5}
}

@misc{Deep2016Alemi,
  author  = {Alemi, A. A. and Fischer, I. and Dillon, J. V. and Murphy, K.},
  title   = {Deep variational information bottleneck},
  howpublished = {arXiv preprint arXiv:1612.00410},
  year    = {2016}
}

@CONFERENCE{How2023Kawaguchi,
  author  = {Kawaguchi, K. and Deng, Z. and Ji, X. and Huang, J.},
  title   = {How does information bottleneck help deep learning?},
  booktitle = {In Proc. Int. Conf. Mach. Learn.}, 
  year    = {2023},
  pages   = {16049-16096}
}

@ARTICLE{Improving2023Kuang,
  author  = {Kuang, H. and Liu, H. and Wu, Y. and Satoh, S. I. and Ji, R.},
  title   = {Improving adversarial robustness via information bottleneck distillation},
  journal = {Adv. Neural Inf. Process. Syst.}, 
  volume  = {36},
  year    = {2023},
  pages   = {10796-10813}
}

@ARTICLE{Gated2023Alesiani,
  author  = {Alesiani, F. and Yu, S. and Yu, X.},
  title   = {Gated information bottleneck for generalization in sequential environments},
  journal = {Knowl. Inf. Syst.}, 
  volume  = {65},
  year    = {2023},
  pages   = {683-705}
}

@ARTICLE{SNIB2023Yang,
  author  = {Yang, S. and Chen, B.},
  title   = {SNIB: improving spike-based machine learning using nonlinear information bottleneck},
  journal = {IEEE Trans. Syst. Man Cybern. Syst.}, 
  volume  = {53},
  year    = {2023},
  pages   = {7852-7863}
}

@ARTICLE{Self2024Yang,
  author  = {Yang, S. and Linares-Barranco, B. and Wu, Y. and Chen, B.},
  title   = {Self-supervised high-order information bottleneck learning of spiking neural network for robust event-based optical flow estimation},
  journal = {IEEE Trans. Pattern Anal. Mach. Intell.}, 
  volume  = {47},
  year    = {2024},
  pages   = {2280-2297}
}

@CONFERENCE{InfoDPCCA2025Tang,
  author  = {Tang, S. and Yu, S.},
  title   = {InfoDPCCA: Information-Theoretic Dynamic Probabilistic Canonical Correlation Analysis},
  howpublished = {In Proc. Conf. Uncertainty Artif. Intell.},
  year    = {2025}
}

@CONFERENCE{Efficient2021Guo,
  author  = {Guo, S. and Lin, T.},
  title   = {An efficient non-backpropagation method for training spiking neural networks},
  booktitle = {In Proc. IEEE Int. Conf. Tools Artif. Intell. (ICTAI)}, 
  year    = {2021},
  pages   = {192-199}
}

@ARTICLE{Information2025Zhang,
  author  = {Zhang, Q. and Lu, M. and Yu, S. and Xin, J. and Chen, B.},
  title   = {An information bottleneck approach for feature selection},
  journal = {Pattern Recognit.},
  volume  = {164},
  year    = {2025},
  pages   = {111564}
}

@ARTICLE{Conditional2020Fischer,
  author  = {Fischer, I.},
  title   = {The conditional entropy bottleneck},
  journal = {Entropy}, 
  volume  = {22},
  year    = {2020},
  pages   = {999}
}

@INCOLLECTION{Elements2011Nalewajski,
  author  = {Nalewajski, R. F.},
  title   = {Elements of information theory},
  booktitle = {Perspectives in Electronic Structure Theory},
  publisher = {Springer Berlin Heidelberg},
  address = {Berlin, Heidelberg},
  year    = {2011},
  pages   = {371-395}
}

@ARTICLE{Diet2021Rathi,
  author  = {Rathi, N. and Roy, K.},
  title   = {Diet-snn: A low-latency spiking neural network with direct input encoding and leakage and threshold optimization},
  journal = {IEEE Trans. Neural Netw. Learn. Syst.}, 
  volume  = {34},
  year    = {2021},
  pages   = {3174-3182}
}

@CONFERENCE{Rmp2020Han,
  author  = {Han, B. and Srinivasan, G. and Roy, K.},
  title   = {Rmp-snn: Residual membrane potential neuron for enabling deeper high-accuracy and low-latency spiking neural network},
  booktitle = {In Proc. IEEE/CVF Conf. Comput. Vis. Pattern Recognit.}, 
  year    = {2020},
  pages   = {13558-13567}
}

@ARTICLE{Snn2022Ding,
  author  = {Ding, J. and Bu, T. and Yu, Z. and Huang, T. and Liu, J.},
  title   = {Snn-rat: Robustness-enhanced spiking neural network through regularized adversarial training},
  journal = {Adv. Neural Inf. Process. Syst.}, 
  volume  = {35},
  year    = {2022},
  pages   = {24780-24793}
}

@misc{Enhancing2024Liu,
  author  = {Liu, Y. and Bu, T. and Ding, J. and Hao, Z. and Huang, T. and Yu, Z.},
  title   = {Enhancing adversarial robustness in SNNs with sparse gradients},
  howpublished = {arXiv preprint arXiv:2405.20355},
  year    = {2024}
}

@ARTICLE{Regularization2023Zhang,
  author  = {Zhang, H. and Cheng, J. and Zhang, J. and Liu, H. and Wei, Z.},
  title   = {A regularization perspective based theoretical analysis for adversarial robustness of deep spiking neural networks},
  journal = {Neural Netw.}, 
  volume  = {165},
  year    = {2023},
  pages   = {164-174}
}

@CONFERENCE{Enhancing2024Ding,
  author  = {Ding, J. and Yu, Z. and Huang, T. and Liu, J. K.},
  title   = {Enhancing the robustness of spiking neural networks with stochastic gating mechanisms},
  booktitle = {In Proc. AAAI Conf. Artif. Intell.}, 
  year    = {2024},
  pages   = {492-502}
}

@ARTICLE{Random2025Wang,
  author  = {Wang, J. and Zhao, D. and Du, C. and He, X. and Zhang, Q. and Zeng, Y.},
  title   = {Random heterogeneous spiking neural network for adversarial defense},
  journal = {iScience}, 
  volume  = {28},
  year    = {2025},
  pages   = {}
}

@ARTICLE{Noise2025Garipova,
  author  = {Garipova, Y. and Yonekura, S. and Kuniyoshi, Y.},
  title   = {Noise and dynamical synapses as optimization tools for spiking neural networks},
  journal = {Entropy}, 
  volume  = {27},
  year    = {2025},
  pages   = {219}
}

@CONFERENCE{Inherent2020Sharmin,
  author  = {Sharmin, S. and Rathi, N. and Panda, P. and Roy, K.},
  title   = {Inherent adversarial robustness of deep spiking neural networks: Effects of discrete input encoding and non-linear activations},
  booktitle = {In Proc. Eur. Conf. Comput. Vis. (ECCV)}, 
  year    = {2020},
  pages   = {399-414}
}

@CONFERENCE{Hire2021Kundu,
  author  = {Kundu, S. and Pedram, M. and Beerel, P. A.},
  title   = {Hire-snn: Harnessing the inherent robustness of energy-efficient deep spiking neural networks by training with crafted input noise},
  booktitle = {In Proc. IEEE/CVF Int. Conf. Comput. Vis. (ICCV)}, 
  year    = {2021},
  pages   = {5209-5218}
}

@ARTICLE{FEEL2024Xu,
  author  = {Xu, M. and Ma, D. and Tang, H. and Zheng, Q. and Pan, G.},
  title   = {FEEL-SNN: Robust spiking neural networks with frequency encoding and evolutionary leak factor},
  journal = {Adv. Neural Inf. Process. Syst.}, 
  volume  = {37},
  year    = {2024},
  pages   = {91930-91950}
}

@misc{Opening2017Shwartz,
  author  = {Shwartz-Ziv, R. and Tishby, N.},
  title   = {Opening the black box of deep neural networks via information},
  howpublished = {arXiv preprint arXiv:1703.00810},
  year    = {2017}
}

@ARTICLE{On2021Geiger,
  author  = {Geiger, B. C.},
  title   = {On information plane analyses of neural network classifiers---A review},
  journal = {IEEE Trans. Neural Netw. Learn. Syst.}, 
  volume  = {33},
  year    = {2021},
  pages   = {7039-7051}
}

@misc{Adaptive2019Chelombiev,
  author  = {Chelombiev, I. and Houghton, C. and O'Donnell, C.},
  title   = {Adaptive estimators show information compression in deep neural networks},
  howpublished = {arXiv preprint arXiv:1902.09037},
  year    = {2019}
}

@ARTICLE{Nonlinear2019Kolchinsky,
  author  = {Kolchinsky, A. and Tracey, B. D. and Wolpert, D. H.},
  title   = {Nonlinear information bottleneck},
  journal = {Entropy}, 
  volume  = {21},
  year    = {2019},
  pages   = {1181}
}

@ARTICLE{Supervised2025Weingarten,
  author  = {Weingarten, N. Z. and Yakhini, Z. and Butman, M. and Bustin, R.},
  title   = {The supervised information bottleneck},
  journal = {Entropy}, 
  volume  = {27},
  year    = {2025},
  pages   = {452}
}

@ARTICLE{Information2021Li,
  author  = {Li, J. and Liu, D.},
  title   = {Information bottleneck theory on convolutional neural networks},
  journal = {Neural Process. Lett.}, 
  volume  = {53},
  year    = {2021},
  pages   = {1385-1400}
}

@misc{Fashion2017Xiao,
  author  = {Xiao, H. and Rasul, K. and Vollgraf, R.},
  title   = {Fashion-mnist: a novel image dataset for benchmarking machine learning algorithms},
  howpublished = {arXiv preprint arXiv:1708.07747},
  year    = {2017}
}

@misc{Learning2009Krizhevsky,
  author  = {Krizhevsky, A. and Hinton, G.},
  title   = {Learning multiple layers of features from tiny images},
  howpublished = {Technical report, University of Toronto},
  year    = {2009}
}

@CONFERENCE{Computing2014Horowitz,
  author  = {Horowitz, M.},
  title   = {1.1 computing's energy problem (and what we can do about it)},
  booktitle = {In Proc. IEEE Int. Solid-State Circuits Conf. (ISSCC)}, 
  year    = {2014},
  pages   = {10-14}
}

@CONFERENCE{Towards2018Madry,
  author  = {Madry, A. and Makelov, A. and Schmidt, L. and Tsipras, D. and Vladu, A.},
  title   = {Towards deep learning models resistant to adversarial attacks},
  booktitle = {In Proc. Int. Conf. Learn. Represent. (ICLR)},
  year    = {2018}
}

@CONFERENCE{Adversarial2017Kurakin,
  author  = {Kurakin, A. and Goodfellow, I. and Bengio, S.},
  title   = {Adversarial examples in the physical world},
  booktitle = {In Proc. Int. Conf. Learn. Represent. Workshop},
  year    = {2017}
}


\end{document}